\documentclass[acmsmall]{acmart}

\AtBeginDocument{%
  \providecommand\BibTeX{{%
    \normalfont B\kern-0.5em{\scshape i\kern-0.25em b}\kern-0.8em\TeX}}}

\usepackage{multirow}
\usepackage{graphicx}
\usepackage{url}
\newcommand{\tabitem}{~~\llap{\textbullet}~~}

\usepackage{color}
\usepackage{soul}

\usepackage{setspace}

\begin{document}

\title{Multi-document Summarization via Deep Learning Techniques: A Survey}

\author{CONGBO MA}
\affiliation{\institution{The University of Adelaide}}
\email{congbo.ma@adelaide.edu.au}

\author{WEI EMMA ZHANG}
\affiliation{\institution{The University of Adelaide}}
\email{wei.e.zhang@adelaide.edu.au}

\author{MINGYU GUO}
\affiliation{\institution{The University of Adelaide}}
\email{mingyu.guo@adelaide.edu.au}

\author{HU WANG}
\affiliation{\institution{The University of Adelaide}}
\email{hu.wang@adelaide.edu.au}

\author{QUAN Z. SHENG}
\affiliation{\institution{Macquarie University}}
\email{michael.sheng@mq.edu.au}

\renewcommand{\shortauthors}{C.Ma et al.}

\setlength{\baselineskip}{11.29 pt}
\begin{abstract}

Multi-document summarization (MDS) is an effective tool for information aggregation that generates an informative and concise summary from a cluster of topic-related documents. Our survey, the first of its kind, systematically overviews the recent deep learning based MDS models. We propose a novel taxonomy to summarize the design strategies of neural networks and conduct a comprehensive summary of the state-of-the-art. We highlight the differences between various objective functions that are rarely discussed in the existing literature. Finally, we propose several future directions pertaining to this new and exciting field.

\end{abstract}



\maketitle

\section{Introduction}

In this era of rapidly advancing technology, the exponential increase of data availability makes analyzing and understanding text files a tedious, labor-intensive, and time-consuming task \cite{oussous2018big, hu2017opinion}. The need to process this abundance of text data rapidly and efficiently calls for new, effective text summarization techniques. Text summarization is a key natural language processing (NLP) tasks that automatically converts a text, or a collection of texts within the same topic, into a concise summary that contains key semantic information which can be beneficial for many downstream applications such as creating news digests, search engine, and report generation \cite{paulus2018deep}.

Text can be summarized from one or several documents, resulting in single document summarization (SDS) and multi-document summarization (MDS). While simpler to perform, SDS may not produce comprehensive summaries because it does not make good use of related, or more recent, documents. Conversely, MDS generates more comprehensive and accurate summaries from documents written at different times, covering different perspectives, but is accordingly more complicated as it tries to resolve potentially diverse and redundant information \cite{tas2007survey}.

In addition, excessively long input documents often lead to model degradation \cite{jin2020multi}. It is challenging for models to retain the most critical contents of complex input sequences while generating a coherent, non-redundant, factual consistent and grammatically readable summary. Therefore, MDS requires models to have stronger capabilities for analyzing the input documents, identifying and merging consistent information.

MDS enjoys a wide range of real-world applications, including summarization of news \cite{fabbri-etal-2019-multi}, scientific publications \cite{yasunaga2019scisummnet}, emails \cite{carenini2007summarizing, zajic2008single}, product reviews \cite{gerani2014abstractive}, medical documents \cite{afantenos2005summarization}, lecture feedback \cite{luo2015summarizing, luo2016automatic}, software project activities \cite{alghamdi2020human}, and Wikipedia articles generation \cite{liu2018generating}. Recently, MDS technology has also received a great amount of industry attention; an intelligent multilingual news reporter bot named Xiaomingbot \cite{xu2020xiaomingbot} was developed for news generation, which can summarize multiple news sources into one article and translate it into multiple languages. Massive application requirements and rapidly growing online data have promoted the development of MDS. 
Existing methods using traditional algorithms are based on: term frequency-inverse document frequency (TF-IDF) \cite{radev2004centroid, baralis2012multi}, clustering \cite {goldstein2000multi, wan2008multi}, graphs \cite{mani1997multi, wan2006improved} and latent semantic analysis \cite{arora2008latent, haghighi2009exploring}.
Most of these works still generate summaries with manually crafted features \cite{mihalcea2005language, wan2006improved}, such as sentence position features \cite{baxendale1958machine, erkan2004lexrank}, sentence length features \cite{erkan2004lexrank}, proper noun features \cite{vodolazova2013extractive}, cue-phrase features \cite{gupta2010survey}, biased word features, sentence-to-sentence cohesion and sentence-to-centroid cohesion. 
Deep learning has gained enormous attention in recent years due to its success in various domains, for instance, computer vision \cite{krizhevsky2012imagenet}, natural language processing \cite{devlin2014fast} and multi-modal learning \cite{hu2020soft}. Both industry and academia have embraced deep learning to solve complex tasks due to its capability of capturing highly nonlinear relations of data. Moreover, deep learning based models reduce dependence on manual feature extraction and pre-knowledge in the field of linguistics, drastically improving the ease of engineering \cite{Amirsina2020natural}. 
Therefore, deep learning based methods demonstrate outstanding performance in MDS tasks in most cases \cite{li2020leveraging,cao2015learning, Lu2020multixscience, liu2019hierarchical, logan2019scoring}. With recently dramatic improvements in computational power and the release of increasing numbers of public datasets, neural networks with deeper layers and more complex structures have been applied in MDS \cite{liu2019hierarchical, li2017cascaded}, accelerating the development of text summarization with more powerful and robust models. These tasks are attracting attention in the natural language processing community; the number of research publications on deep learning based MDS has increased rapidly over the last five years.

\begin{figure}[t]
\begin{center}
\includegraphics[width=1\textwidth]{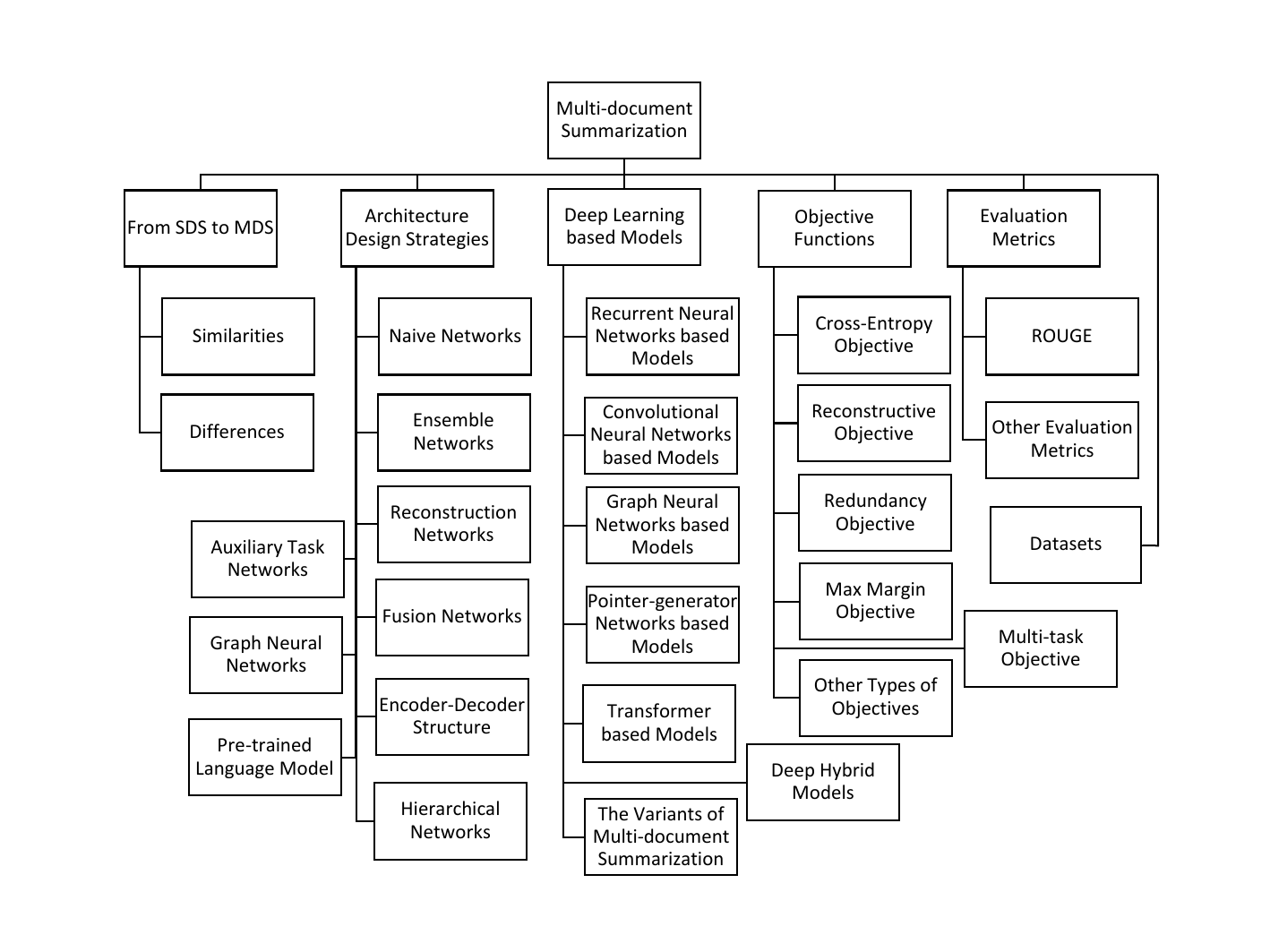}
\end{center}
\vspace{-10mm}
\caption{Hierarchical Structure of This Survey. }
\vspace{-2mm}
\label{fig:structure}
\end{figure}

The prosperity of deep learning for summarization in both academia and industry requires a comprehensive review of current publications for researchers to better understand the process and research progress. However, most of the existing summarization survey papers are based on traditional algorithms instead of deep learning based methods or target general text summarization  \cite{nenkova2012survey,haque2013literature,ferreira2014multi,shah2016literature,eswa/El-KassasSRM21}. We have therefore surveyed recent publications on deep learning methods for MDS that, to the best of our knowledge, is the first comprehensive survey of this field.
This survey has been designed to classify neural based MDS techniques into diverse categories thoroughly and systematically. We also conduct a detailed discussion on the categorization and progress of these approaches to establish a clearer concept standing in the shoes of readers. We hope this survey provides a panorama for researchers, practitioners and educators to quickly understand and step into the field of deep learning based MDS. The key contributions of this survey are three-fold:
\begin{itemize}
\item We propose a categorization scheme to organize current research and provide a comprehensive review for deep learning based MDS techniques, including deep learning based models, objective functions, benchmark datasets and evaluation metrics. 

\item We review development movements and provide a systematic overview and summary of the state-of-the-art. We also summarize nine network design strategies based on our extensive studies of the current models.

\item We discuss the open issues of deep learning based multi-document summarization and identify the future research directions of this field. We also propose potential solutions for some discussed research directions.
\end{itemize}

\vspace{1mm}
\noindent \textbf{Paper Selection.}   
We used Google Scholar as the main search engine to select representative works from 2015 to 2021. High-quality papers were selected from top NLP and AI journals and conferences, include  ACL\footnote{Annual Meeting of the Association for Computational Linguistics.}, EMNLP\footnote{Empirical Methods in Natural Language Processing.}, COLING\footnote{International Conference on Computational Linguistics}, NAACL\footnote{Annual Conference of the North American Chapter of the Association for Computational Linguistics.}, AAAI\footnote{AAAI Conference on Artificial Intelligence.}, ICML\footnote{International Conference on Machine Learning.}, ICLR\footnote{International Conference on Learning Representations} and IJCAI\footnote{International Joint Conference on Artificial Intelligence.}. The major keywords we used include \textit{multi-documentation summarization}, \textit{summarization}, \textit{extractive summarization}, \textit{abstractive summarization}, \textit{deep learning} and \textit{neural networks}.

\vspace{2mm}
\noindent \textbf{Organization of the Survey.}   
This survey will cover various aspects of recent advanced deep learning based works in MDS. Our proposed taxonomy categorizes the works from six aspects (Figure \ref{fig:structure}). To be more self-contained, in Section \ref{sec:overview}, we give the problem definition, the processing framework of text summarization, discuss similarities and differences of between SDS and MDS. Nine deep learning architecture design strategies, six deep learning based methods, and the variant tasks of MDS are presented in Section \ref{sec: methods}. Section \ref{sec: objective} summarizes objective functions that guide the model optimization process in the reviewed literature while evaluation metrics in Section \ref{sec: evaluation} help readers choose suitable indices to evaluate the effectiveness of a model. Section \ref{sec: datasets} summarizes standard and the variant MDS datasets. Finally, Section \ref{sec: futurework} discusses future research directions for deep learning based MDS followed by conclusions in Section \ref{sec: conclusion}.

\begin{figure}[t]
\begin{center}
\includegraphics[width=0.95\textwidth]{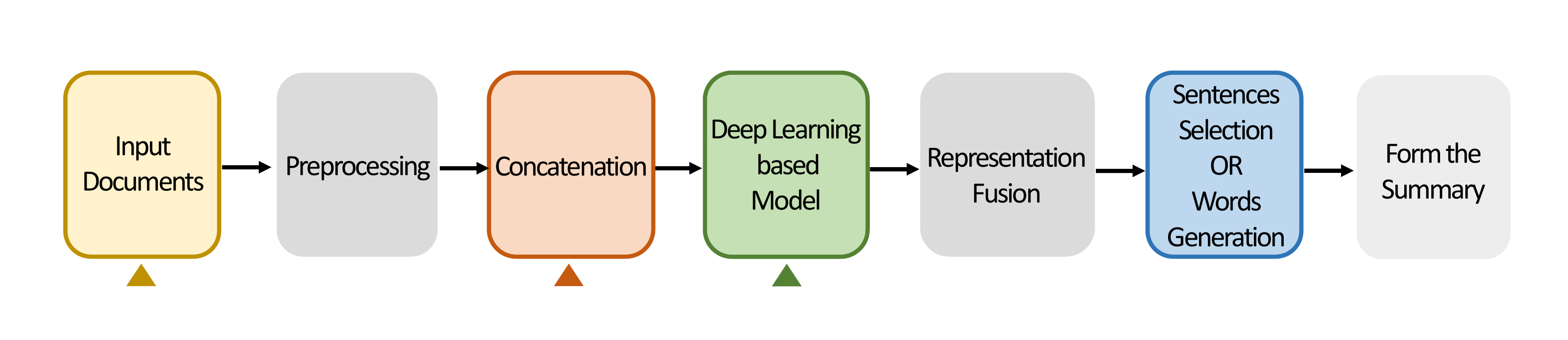}
\end{center}
\vspace{-5mm}
\caption{The Processing Framework of Text Summarization. Each of the highlighted steps (the one with triangle mark) indicates the differences between SDS and MDS. }
\vspace{-3mm}
\label{fig:framework}
\end{figure}

\section{From Single to Multi-document Summarization}
\label{sec:overview}

Before we dive into the details of existing deep learning based techniques, we start by defining SDS and MDS, and introducing the concepts used un both methods. The aim of MDS is to generate a concise and informative summary $Sum$ from a collection of documents $D$. $D$ denotes a cluster of topic-related documents $\left \{ d_{i}\mid i\in [1, N] \right \}$, where $N$ is the number of documents. Each document $d_{i}$ consists of $M_{d_{i}}$ sentences $\left \{ s_{i,j}\mid j\in [1, M_{d_{i}}] \right \}$.  $s_{i,j}$  refers to the $j$-th sentence in the $i$-th document. The standard summary $Ref$ is called the \textsl{golden summary} or \textsl{reference summary}. Currently, most golden summaries are written by experts. We keep this notation consistent throughout the article.  

To give readers a clear understanding of the processing of deep learning based summarization tasks, we summarize and illustrate the processing framework as shown in Figure \ref{fig:framework}. The first step is preprocessing input document(s), such as segmenting sentences, tokenizing non-alphabetic characters, and removing punctuation \cite{shirwandkar2018extractive}. MDS models in particular need to select suitable concatenation methods to capture cross-document relations. Then, an appropriate deep learning based model is chosen to generate semantic-rich representation for downstream tasks. The next step is to fuse these various types of representation for later sentence selection or summary generation. Finally, document(s) are transformed into a concise and informative summary. Each of the highlighted steps in Figure \ref{fig:framework} (indicated by triangles) indicates a difference between SDS and MDS. Based on this process, the research questions of MDS can be summarized as follows:

\begin{itemize}
\item How to capture the cross-document relations and in-document relations from the input documents? 
\item Compared to SDS, how to extract or generate salient information in a larger search space containing conflict, duplication and complementary information?
\item How to best fuse various representation from deep learning based models and external knowledge?
\item How to comprehensively evaluate the performance of MDS models? 
\end{itemize}

The following sections provide a comprehensive analysis of the similarities and differences between SDS and MDS.

\begin{figure}[t]
\begin{center}
\includegraphics[width=1\textwidth]{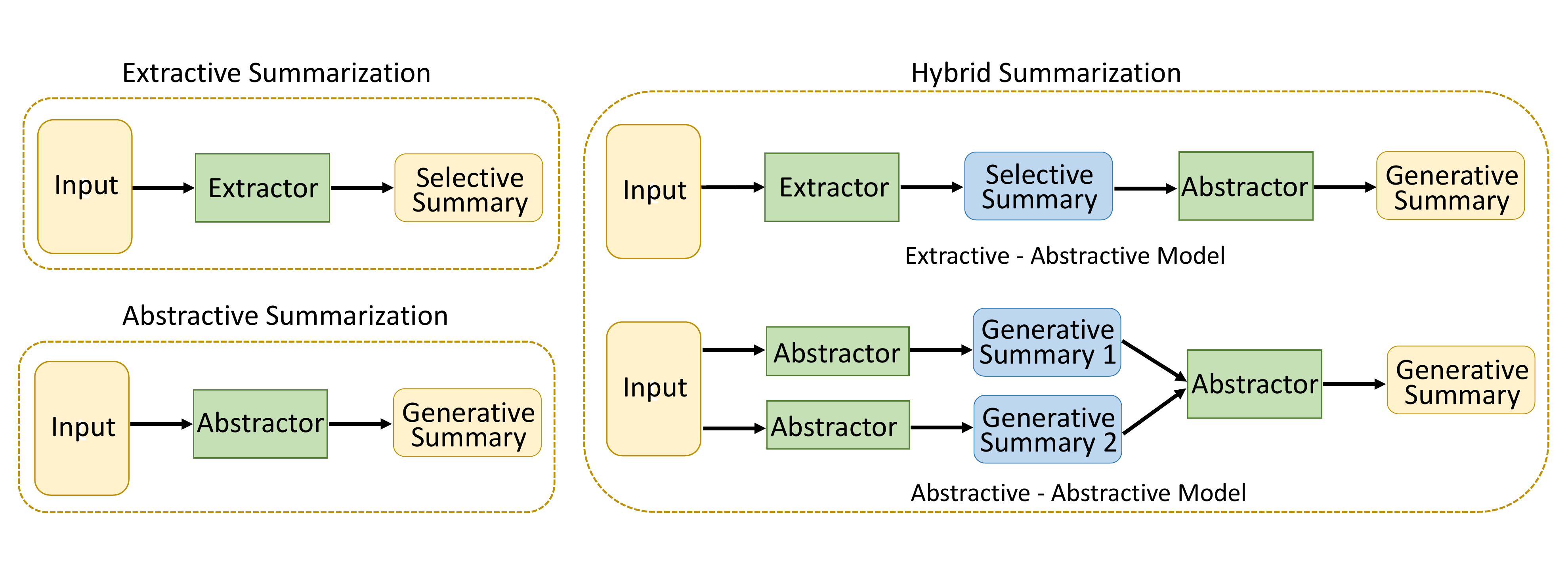}
\end{center}
\vspace{-8mm}
\caption{Summarization Construction Types for Text Summarization.}\label{fig:summarization construction types}
\vspace{-2mm}
\end{figure}

\subsection{Similarities between SDS and MDS}
Existing SDS and MDS methods share the summarization construction types, learning strategies, evaluation indexes and objective functions. SDS and MDS both seek to compress the document(s) into a short and informative summary. Existing summarization methods can be grouped into \textsl{abstractive summarization}, \textsl{extractive summarization} and \textsl{hybrid summarization} (Figure \ref{fig:summarization construction types}). Extractive summarization methods select salient snippets from the source documents to creat informative summaries, and generally contain two major components: \textsl{sentence ranking} and \textsl{sentence selection} \cite{cao2015ranking, nallapati2017summarunner}. Abstractive summarization methods aim to present the main information of input documents by automatically generating summaries that are both succinct and coherent; this cluster of methods allows models to generate new words and sentences from a corpus pool \cite{paulus2018deep}. Hybrid models are proposed to combine the advantages of both extractive and abstractive methods to process the input texts. Research on summarization focuses on two learning strategies. One strategy seeks to enhance the generalization performance by improving the architecture design of the end-to-end models \cite{fabbri-etal-2019-multi, chu2019meansum, jin2020multi, liu2019hierarchical}. The other leverages external knowledge or other auxiliary tasks to complement summary selection or generation \cite{cao2017improving, li2020leveraging}. Furthermore, both SDS and MDS aim to minimize the distance between machine-generated summary and golden summary. Therefore, SDS and MDS could share some indices to evaluate the performance of summarization models such as Recall-Oriented Understudy for Gisting Evaluation (ROUGE, see Section \ref{sec: evaluation}), and objective functions to guide model optimization.

\subsection{Differences between SDS and MDS}
In the early stages of MDS, researchers directly applied SDS models to MDS \cite{mao2020multi}. However, a number of aspects in MDS that are different from SDS and these differences are also the breakthrough point for exploring the MDS models. We summarize the differences in the following five aspects: 

\begin{itemize}
\item More diverse input document types;
\item  Insufficient methods to capture cross-document relations;
\item High redundancy and contradiction across input documents;
\item  Larger searching space but lack of sufficient training data;
\item  Lack of evaluation metrics specifically designed for MDS.
\end{itemize}

A defining different character between SDS and MDS is the number of input documents. The MDS tasks deal with multiple sources, of types that can be roughly divided into three groups:

\begin{itemize}

\item Many short sources, where each document is relatively short but the quantity of the input data is large. A typical example is  product reviews summarization that aims to generate a short, informative summary from numerous individual reviews \cite{angelidis2018summarizing}.
\item  Few long sources. For example, generating a summary from a group of news articles \cite{fabbri-etal-2019-multi}, or constructing a Wikipedia style article from several web articles \cite{liu2018generating}. 
\item  Hybrid sources containing one or few long documents with several to many shorter documents. For example, news article(s) with several readers' comments to this news \cite{li2017reader}, or a scientific summary from a long paper with several short corresponding citations \cite{yasunaga2019scisummnet}.
\end{itemize}

As SDS only uses one input document, no additional processing is required to assess relationships between SDS inputs. By their very nature, the multiple input documents used in MDS are likely to contain more contradictory, redundant, and complementary information \cite{radev2000common}. MDS models therefore require sophisticated algorithms to identify and cope with redundancy and contradictions across documents to ensure that the final summary is comprehensive. Detecting these relations across documents can bring benefits for MDS models. In the MDS tasks, there are two common methods to concatenate multiple input documents:
\begin{figure}[t]
\begin{center}
\includegraphics[width=0.76\textwidth]{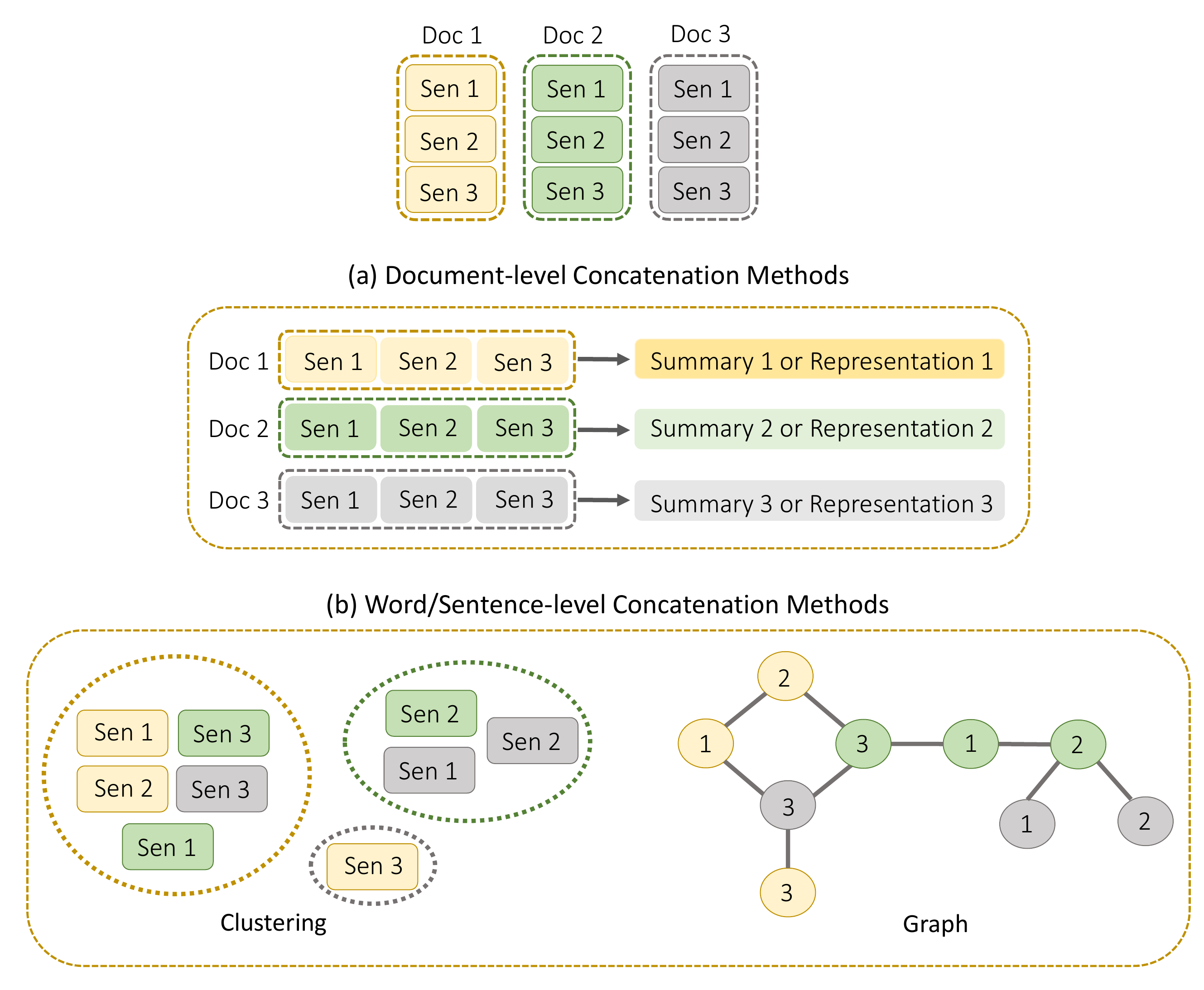}
\vspace{-2.5mm}
\end{center}
\caption{The Methods of Hierarchical Concatenation.}\label{fig:Hierarchical concatenate}
\vspace{-3mm}
\end{figure}

\begin{itemize}
   
\item Flat concatenation is a simple yet powerful concatenation method, where all input documents are spanned and processed as a flat sequence; to a certain extent, this method converts MDS to an SDS tasks. Inputting flat-concatenated documents requires models to have strong ability to process long sequences.
\item Hierarchical concatenation is able to preserve cross-document relations. However, many existing deep learning methods do not make full use of this hierarchical relationship \cite{wang2020heterogeneous, fabbri-etal-2019-multi, liu2018generating}. Taking advantage of hierarchical relations among documents instead of simply flat concatenating articles facilitates the MDS model to obtain representation with built-in hierarchical information, which in turn improves the effectiveness of the models.
The input documents within a cluster describe a similar topic logically and semantically. Figure \ref{fig:Hierarchical concatenate} illustrates two representative methods of hierarchical concatenation. Existing hierarchical concatenation methods either perform document-level condensing in a cluster separately \cite{amplayo2019informative} or process documents in word/sentence-level inside document cluster \cite{mir2018abstractive, antognini2019learning, wang2020heterogeneous}. 
In Figure \ref{fig:Hierarchical concatenate}(a), the extractive or abstractive summaries, or representation from the input documents are fused in the subsequent processes for final summary generation. The models using document-level concatenation methods are usually two-stage models.
In Figure \ref{fig:Hierarchical concatenate}(b), sentences in the documents can be replaced by words. For the word or sentence-level concatenation methods, clustering algorithms and graph-based techniques are the most commonly used methods. Clustering methods could help MDS models decrease redundancy and increase the information coverage for the generated summaries \cite{mir2018abstractive}.
Sentence relation graph is able to model hierarchical relations among multi-documents as well \cite{antognini2019learning, yasunaga2019scisummnet, yasunaga2017graph}. Most of the graph construction methods utilize sentences as vertexes and the edge between two sentences indicates their sentence-level relations \cite{antognini2019learning}. Cosine similarity graph \cite{erkan2004lexrank}, discourse graph \cite{christensen2013towards, yasunaga2017graph, liu2019hierarchical}, semantic graph \cite{pasunuru2021efficiently} and heterogeneous graph \cite{wang2020heterogeneous} can be used for building sentence graph structures. These graph structure could all serve as external knowledge to improve the performance of MDS models. 
\end{itemize}

In addition to capture cross-document relation, hybrid summarization models can also be used to capture complex documents semantically, as well as to fuse disparate features that are more commonly adopted by MDS tasks. These models usually process data in two stages: extractive-abstractive and abstractive-abstractive (the right part of Figure \ref{fig:summarization construction types}). The two-stage models try to gather important information from source documents with extractive or abstractive methods at the first stage, to significantly reduce the length of documents. In the second stage, the processed texts are fed into an abstractive model to form final summaries \cite{amplayo2019informative, logan2019scoring, liu2018generating, liu2019hierarchical, li2020leveraging}.

Furthermore, conflict, duplication, and complementarity among multiple source documents require MDS models to have stronger abilities to handle complex information. However, applying the SDS model directly on MDS tasks is difficult to handle much higher redundancy \cite{mao2020multi}. Therefore, the MDS models are required not only to generate coherent and complete summary but also more sophisticated algorithms to identify and cope with redundancy and contradictions across documents ensuring that the final summary should be complete in itself. MDS also involves larger searching spaces but has smaller-scale training data than SDS, which sets obstacles for deep learning based models to learn adequate representation \cite{mao2020multi}. In addition, there are no specific evaluation metrics designed for MDS; however, existing SDS evaluation metrics can not evaluate the relationship between the generated abstract and different input documents well.

\vspace{1mm}
\section{Deep Learning Based Multi-document Summarization Methods} \label{sec: methods}

Deep neural network (DNN) models learn multiple levels of representation and abstraction from input data and can fit data in a variety of research fields, such as computer vision \cite{krizhevsky2012imagenet} and natural language process \cite{devlin2014fast}. Deep learning algorithms replace manual feature engineering by learning distinctive features through back-propagation to minimize a given objective function. It is well known that linear solvable problems possess many advantages, such as being easily solved and having numerous theoretically proven supports; however, many NLP tasks are highly non-linear. As theoretically proven by Hornik et al. \cite{hornik1989multilayer}, neural networks can fit any given continuous function as a universal approximator. For the MDS tasks, DNNs also perform considerably better than traditional methods to effectively process large-scale documents and distill informative summaries due to their strong fitting abilities. In this section, we first introduce our novel taxonomy that generalizes nine neural network design strategies (Section 3.1). We then present the state-of-the-art DNN based MDS models according to the main neural network architecture they adopt (Section 3.2 – 3.7), before finishing with a  brief introduction to MDS variant tasks (Section 3.8).

\subsection{Architecture Design Strategies}

Architecture design strategies play a critical role in deep learning based models, and many architectures have been applied to variants MDS tasks. Here, we have generalized the network architectures and summarize them into nine types based on how they generate or fuse semantic-rich and syntactic-rich representation to improve MDS model performance (Figure \ref{fig:network}); these different architectures can also be used as basic structures or stacked on each other to obtain more diverse design strategies. In Figure \ref{fig:network}, deep neural models are in green boxes, and can be flexibly substituted with other backbone networks. The blue boxes indicate the neural embeddings processed by neural networks or heuristic-designed approaches, e.g., "sentence/document" or "other" representation. The explanation of each sub-figure is listed as follows:

\begin{figure}[t]
\begin{center}
\includegraphics[width=1.05\textwidth]{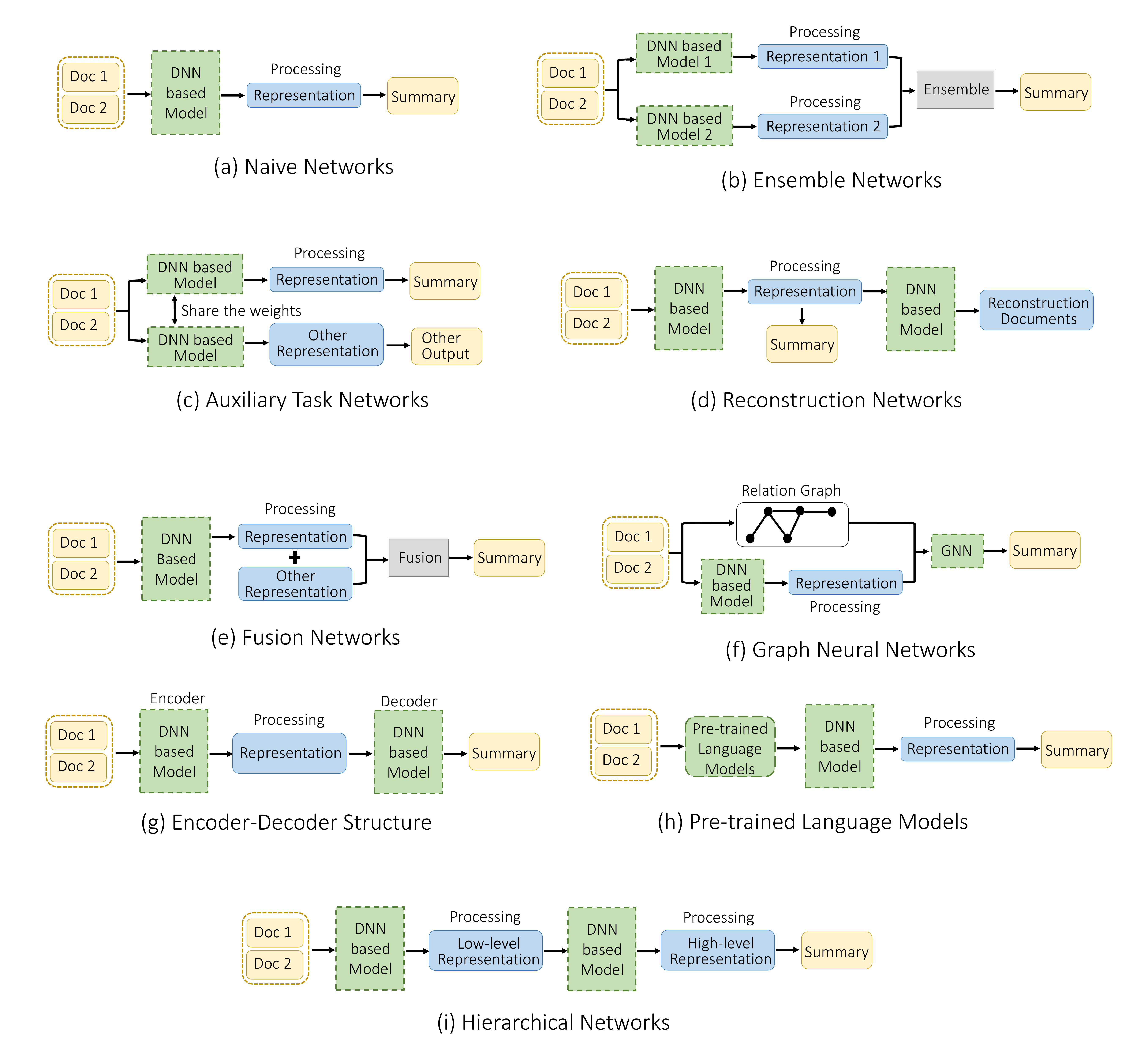}
\end{center}
\vspace{-4mm}
\caption{ Network Design Strategies.}\label{fig:network}
\vspace{-2mm}
\end{figure}

\begin{itemize}
\item \textsl{Naive Networks (Figure \ref{fig:network}(a)). }   
Multiple concatenated documents are input through DNN based models to extract features. Word-level, sentence-level or document-level representation is used to generate the downstream summary or select sentences. Naive networks represent the most naive model that lays the foundation for other strategies.

\item \textsl{Ensemble Networks (Figure \ref{fig:network}(b)).}
Ensemble based methods leverage multiple learning algorithms to obtain better performance than individual algorithms. To capture semantic-rich and syntactic-rich representation, Ensemble networks feed input documents to multiple paths with different network structures or operations. Later on, the representation from different networks is fused to enhance model expression capability. The majority vote or the average score can be used to determine the final output.

\item \textsl{Auxiliary Task Networks (Figure \ref{fig:network}(c))} 
employ different tasks in the  summarization models, where text classification, text reconstruction or other auxiliary tasks serve as complementary representation learners to obtain advanced features.  Meanwhile, auxiliary task networks also provide researchers with a solution to use appropriate data from other tasks. In this strategy, parameters sharing scheme are used for jointly optimizing different tasks. 

\item \textsl{Reconstruction Networks (Figure \ref{fig:network}(d))}
optimize models from an unsupervised learning paradigm, which allows summarization models to overcome the limitation of insufficient annotated golden summaries. The use of such a paradigm enables generated summaries to be constrained in the natural language domain in a good manner.

\item \textsl{Fusion Networks (Figure \ref{fig:network}(e))}
fuse representation generated from neural networks and hand-crafted features. These hand-crafted features contain adequate prior knowledge that facilitates the optimization of summarization models.

\item \textsl{Graph Neural Networks (Figure \ref{fig:network}(f)).}
This strategy captures cross-document relations, crucial and beneficial for multi-document model training, by constructing graph structures based on the source documents, including word, sentence, or document-level information.

\item \textsl{Encoder-Decoder Structure (Figure \ref{fig:network}(g)).} The encoder embeds source documents into the hidden representation, i.e., word, sentence and document representation. This representation, containing compressed semantic and syntactic information, is passed to the decoder which processes the latent embeddings to synthesize local and global semantic/syntactic information to produce the final summaries.

\item \textsl{Pre-trained Language Models (Figure \ref{fig:network}(h))} obtain contextualized text representation by predicting words or phrases based on their context using large amounts of the corpus, which can be further fine-tuned for downstream task adaption \cite{dong2019unified}. The models can fine-tune with randomly initialized decoders in an end-to-end fashion since transfer learning can assist the model training process \cite{li2020leveraging}.

\item \textsl{Hierarchical Networks (Figure \ref{fig:network}(i)).}
Multiple documents are concatenated as inputs to feed into the first DNN based model to capture low-level representation. Another DNN based model is cascaded to generate high-level representation based on the previous ones. The hierarchical networks empower the model with the ability to capture abstract-level and semantic-level features more efficiently.
\end{itemize}

\vspace{-1mm}
\subsection{Recurrent Neural Networks based Models}
 
Recurrent Neural Networks (RNNs) \cite{rumelhart1986learning} excel in modeling sequential data by capturing sequential relations and syntactic/semantic information from word sequences. In RNN models, neurons are connected through hidden layers and unlike other neural network structures, the inputs of each RNN neuron come not only from the word or sentence embedding but also from the output of the previous hidden state. Despite being powerful, vanilla RNN models often encounter gradient explosion or vanishing issues, so a large number of RNN-variants have been proposed. The most prevalent ones are Long Short-Term Memory (LSTM) \cite{hochreiter1997long},  Gated Recurrent Unit (GRU)  \cite{Chung2014empirical} and Bi-directional Long Short-Term Memory (Bi-LSTM) \cite{huang2015bidirectional}. The DNN based Model in Figure \ref{fig:network} can be replaced with RNN based models to design models.

RNN based models have been used in MDS tasks since 2015. Cao et al. \cite{cao2015ranking} proposed an RNN-based model termed \textsl {Ranking framework upon Recursive Neural Networks (R2N2)}, which leverages manually extracted words and sentence-level features as inputs. This model transfers the sentence ranking task into a hierarchical regression process, which measures the importance of sentences and constituents in the parsing tree. Zheng et al. \cite{zheng2019subtopic} used a hierarchical RNN structure to utilize the subtopic information by extracting not only sentence and document embeddings, but also topic embeddings. In this SubTopic-Driven Summarization \textsl {(STDS)} model, the readers' comments are seen as auxiliary documents and the model employs soft clustering to incorporate comment and sentence representation for further obtaining subtopic representation.
Arthur et al. \cite{bravzinskas2019unsupervised} introduced a GRU-based encoder-decoder architecture to minimize the diversity of opinions reflecting the dominant views while generating multi-review summaries. Mao et al. \cite{mao2020multi} proposed a maximal margin relevance guided  reinforcement learning framework (RL-MMR) to incorporate the advantages of neural sequence learning and statistical measures. The proposed soft attention for learning adequate representation allows more exploration of search space. 
 
To leverage the advantage of hybrid summarization model, Reinald et al. \cite{amplayo2019informative} proposed a two-stage framework, viewing opinion summarization as an instance of multi-source transduction to distill salient information from source documents. The first stage of the model leverages a Bi-LSTM auto-encoder to learn word and document-level representation; the second stage fuses multi-source representation and generates an opinion summary with a simple LSTM decoder combined with a vanilla attention mechanism \cite{bahdanau2015neural} and a copy mechanism \cite{oriol2017pointer}.

Since paired MDS datasets are rare and hard to obtain, Li et al. \cite{li2017cascaded} developed a RNN-based framework to extract salient information vectors from sentences in input documents in an unsupervised manner. Cascaded attention retains the most relevant embeddings to reconstruct the original input sentence vectors. During the reconstruction process, the proposed model leverages a sparsity constraint to penalize trivial information in the output vectors. Also, Chu et al. \cite{chu2019meansum} proposed an unsupervised end-to-end abstractive summarization architecture called \textsl {MeanSum}. This LSTM-based model formalizes product or business reviews summarization problem into two individual closed-loops. 
Inspired by MeanSum, Coavoux et al. \cite{coavoux2019unsupervised} used a two-layer standard LSTM to construct sentence representation for aspect-based multi-document abstractive summarization, and discovered that the clustering strategy empowers the model to reward review diversity and handle contradictory ones.

\vspace{-0.3mm}
\subsection{Convolutional Neural Networks Based Models}

Convolutional neural networks (CNNs) \cite{lecun1998gradient} achieve excellent results in computer vision tasks. The convolution operation scans through the word/sentence embeddings and uses convolution kernels to extract important information from input data objects. Using a pooling operation at intervals can return simple to complex feature levels. CNNs have been proven to be effective for various NLP tasks in recent years \cite{kim2014convolutional, dos2014deep} as they can process natural language after sentence/word vectorization. Most of the CNN based MDS models use CNNs for semantic and syntactic feature representation. As with RNN, CNN-based models can also replace DNN-based models in network design strategies (Please refer to Figure \ref{fig:network}).

A simple way to use CNNs in MDS is by sliding multiple filters with different window sizes over the input documents for semantic representation. Cao et al. \cite{cao2015learning} proposed a hybrid CNN-based model \textsl {PriorSum} to capture latent document representation. The proposed representation learner slides over the input documents with filters of different window widths and two-layer max-over-time pooling operations \cite{collobert2011natural} to fetch document-independent features that are more informative than using standard CNNs. 

Similarly, \textsl {HNet} \cite{singh2018unity} uses distinct CNN filters and max-over-time-pooling to generate salient feature representation for downstream processes. Cho et al. \cite{cho2019improving} also used different filter sizes in \textsl{DPP-combined} model to extract low-level features. Yin et al. \cite{yin2015optimizing} presented an unsupervised CNN-based model termed \textsl {Novel Neural Language Model (NNLM)} to extract sentence representation and diminish the redundancy of sentence selection. The NNLM framework contains only one convolution layer and one max-pooling layer, and both element-wise averaging sentence representation and context words representation are used to predict the next word. For aspect-based opinion summarization, Stefanos et al. \cite{angelidis2018summarizing} leveraged a CNN based model to encode the product reviews which contain a set of segments for opinion polarity. 

People with different background knowledge and understanding can produce different summaries of the same documents. To account for this variability, Zhang et al. \cite{zhang2016multiview} suggested a \textsl {MV-CNN} model that ensembles three individual models to incorporate multi-view learning and CNNs to improve the performance of MDS. In this work, three CNNs with dual-convolutional layers used multiple filters with different window sizes to extract distinct saliency scores of sentences.

To overcome the MDS bottlenecks of insufficient training data, Cao et al. \cite{cao2017improving} developed a \textsl{TCSum} model incorporating an auxiliary text classification sub-task into MDS to introduce more supervision signals. The text classification model uses a CNN descriptor to project documents onto the distributed representation, and to classify input documents into different categories. The summarization model shares the projected sentence embedding from the classification model, and the TCSum model then chooses the corresponding category based transformation matrices according to classification results to transform the sentence embedding into the summary embedding.

Unlike RNNs that support the processing of long time-serial signals, a naive CNN layer struggles to capture long-distance relations while processing sequential data due to the limitation of the fixed-sized convolutional kernels, each of which has a specific receptive field size. Nevertheless, CNN based models can increase their receptive fields through formation of hierarchical structures to calculate sequential data in a parallel manner. Because of this highly parallelizable characteristic, training of CNN-based summarization models is more efficient than for RNN-based models. However, summarizing lengthy input articles is still a challenging task for CNN based models because they are not skilled in modeling non-local relationships.

\subsection{Graph Neural Networks Based Models}
CNNs have been successfully applied to many computer vision tasks to extract distinguished image features from the Euclidean space, but struggle when processing non-Euclidean data. Natural language data consist of vocabularies and phrases with strong relations which can be better represented with graphs than with sequential orders. Graph neural networks (GNNs, Figure \ref{fig:network} (f)) are composed of an ideal architecture for NLP since they can model strong relations between entities semantically and syntactically. Graph convolution networks (GCNs) and graph attention networks (GANs) are the most commonly adopted GNNs because of their efficiency and simplicity for integration with other neural networks. These models first build a relation graph based on input documents, where nodes can be words, sentences or documents, and edges capture the similarity among them. At the same time, input documents are fed into a DNN based model to generate embeddings at different levels. The GNNs are then built over the top to capture salient contextual information. Table \ref{tab:graph_based_models} describes the current GNN based models used for MDS with details of nodes, edges, edge weights, and applied GNN methods.

\begin{table}[t]
\caption{Multi-document Summarization Models based on Graph Neural Networks.}
\vspace{-1mm}
\label{tab:graph_based_models}
\scalebox{0.8}{
\begin{tabular}{|c|c|c|c|c|}
\hline
\textbf{Models}                                                               & \textbf{Nodes}                                                                  & \textbf{Edges}                                                                      & \textbf{Edge Weights}                                                               & \textbf{GNN Methods}                                                                    \\ \hline
\begin{tabular}[c]{@{}c@{}} \textsl {HeterDoc-}\\    \textsl{SumGraph} \cite{wang2020heterogeneous}\end{tabular}     & \begin{tabular}[c]{@{}c@{}}word, sentence,\\    document\end{tabular} & \begin{tabular}[c]{@{}c@{}}word-sentence,\\    word-document\end{tabular} & TF-IDF                                                                    & \begin{tabular}[c]{@{}c@{}}Graph Attention\\    Networks\end{tabular}     \\ \hline
\begin{tabular}[c]{@{}c@{}}\textsl{Graph-based}\\    \textsl{Neural MDS} \cite{yasunaga2017graph}\end{tabular} & sentence                                                              & sentence-sentence                                                         & \begin{tabular}[c]{@{}c@{}}Personalized\\    Discourse Graph\end{tabular} & \begin{tabular}[c]{@{}c@{}}Graph Convolutional\\    Networks\end{tabular} \\ \hline
\textsl{SemSentSum} \cite{antognini2019learning}                                                          & sentence                                                              & sentence-sentence                                                         & \begin{tabular}[c]{@{}c@{}}Cosine Similarity Graph    \\ Edge Removal Method  \end{tabular}      & \begin{tabular}[c]{@{}c@{}}Graph Convolutional\\    Networks\end{tabular} \\ \hline
\textsl{ScisummNet} \cite{yasunaga2019scisummnet}                                                          & sentence                                                              & sentence-sentence                                                         & \begin{tabular}[c]{@{}c@{}}Cosine Similarity Graph\end{tabular}      & \begin{tabular}[c]{@{}c@{}}Graph Convolutional\\    Networks\end{tabular} \\ \hline
\end{tabular}} 
\vspace{-2mm}
\end{table}

Yasunage et al. \cite{yasunaga2017graph} developed a GCN based extractive model to capture the relations between sentences. This model first builds a sentence-based graph and then feeds the pre-processed data into a GCN \cite{kipf2017semi} to capture sentence-wise related features. Defined by the model, each sentence is regarded as a node and the relation between each pair of sentences is defined as an edge. Inside each document cluster, the sentence relation graph can be generated through a cosine similarity graph \cite{erkan2004lexrank}, approximate discourse graph \cite{christensen2013towards}, and the proposed personalized discourse graph. Both the sentence relation graph and sentence embeddings extracted by a sentence-level RNN are fed into GCN to produce the final sentence representation. With the help of a document-level GRU, the model generates cluster embeddings to fully aggregate features between sentences. 

Similarly, Antognini et al. \cite{antognini2019learning} proposed a GCN based model named \textsl{SemSentSum} that constructs a graph based on sentence relations. In contrast to Yasunage et al. \cite{yasunaga2017graph}, this work leverages external universal embeddings, pre-trained on the unrelated corpus, to construct a sentence semantic relation graph. Additionally, an edge removal method has been applied to deal with the sparse graph problems emphasizing high sentence similarities; if the weight of the edge is lower than a given threshold, the edge is removed. The sentence relation graph and sentence embeddings are fed into a GCN \cite{kipf2017semi} to generate saliency estimation for extractive summaries. 

Yasunage et al. \cite{yasunaga2019scisummnet} also designed a GCN based model for summarizing scientific papers. The proposed \textsl{ScisummNet} model uses not only the abstract of source scientific papers but also the relevant text from papers that cite the original source. The total number of citations is also incorporated in the model as an authority feature. A cosine similarity graph is applied to form the sentence relation graph, and GCNs are adopted to predict the sentence salience estimation from the sentence relation graph, authority scores and sentence embeddings.

Existing GNN based models focused mainly on the relationships between sentences, and do not fully consider the relationships between words, sentences, and documents. To fill this gap, Wang et al. \cite{wang2020heterogeneous} proposed a heterogeneous GAN based model, called \textsl{HeterDoc-SUM Graph}, that is specific for extractive MDS. This heterogeneous graph structure includes word, sentence, and document nodes, where sentence nodes and document nodes are connected according to the contained word nodes. Word nodes thus act as an intermediate bridge to connect the sentence and document nodes, and are used to better establish document-document, sentence-sentence and sentence-document relations. TF-IDF values are used to weight word-sentence and word-document edges, and the node representation of these three levels are passed into the graph attention networks for model update. In each iteration, bi-directional updating of both word-sentence and word-document relations are performed to better aggregate cross-level semantic knowledge.

\subsection{Pointer-generator Networks Based Models}
Pointer-generator (PG) networks \cite{abigail2017get} are proposed to overcome the problems of factual errors and high redundancy in the summarization tasks. This network has been inspired by Pointer Network \cite{oriol2017pointer}, CopyNet \cite{gu2016incorporating}, forced-attention sentence compression \cite{miao2016language}, and coverage mechanism from machine translation \cite{tu2016modeling}. PG networks combine sequence-to-sequence (Seq2Seq) model and pointer networks to obtain a united probability distribution allowing vocabularies to be selected from source texts or generated by machines. Additionally, the coverage mechanism prevents PG networks from consistently choosing the same phrases.

The \textsl{Maximal Marginal Relevance (MMR)} method is designed to select a set of salient sentences from source documents by considering both \textsl{importance} and \textsl{redundancy} indices \cite{Jaime1998the}. The redundancy score controls sentence selection to minimize overlap with the existing summary. The MMR model adds a new sentence to the objective summary based on importance and redundancy scores until the summary length reaches a certain threshold. Inspired by MMR, Alexander et al. \cite{fabbri-etal-2019-multi} proposed an end-to-end \textsl{Hierarchical MMR-Attention Pointer-generator (Hi-MAP)} model to incorporate PG networks and MMR \cite{Jaime1998the} for abstractive MDS.  The Hi-MAP model improves PG networks by modifying attention weights (multipling MMR scores by the original attention weights) to include better important sentences in, and filter redundant information from, the summary.
Similarly, the MMR approach is implemented by \textsl{PG-MMR} model \cite{logan2018adapting} to identify salient source sentences from multi-document inputs, albeit with a different method for calculating MMR scores from Hi-MAP; instead, ROUGE-L Recall and ROUGE-L Precision \cite{lin2004rouge} serve as evaluation metrics to calculate the importance and redundancy scores. To overcome the scarcity of MDS datasets, the PG-MMR model leverages a support vector regression model that is pre-trained on a SDS dataset to recognize the important contents. This support vector regression model also calculates the score of each input sentence by considering four factors: sentence length, sentence relative/absolute position, sentence-document similarities, and sentence quality obtained by a PG network. Sentences with the top-$K$ scores are fed into another PG network to generate a concise summary.

\subsection{Transformer Based Models}
As discussed, CNN based models are not as good at processing sequential data as RNN based models. However, RNN based models are not amenable to parallel computing, as the current states in RNN models highly depend on results from the previous steps. Additionally, RNNs struggle to process long sequences since former knowledge will fade away during the learning process. Adopting \textsl{Transformer} based architectures \cite{vaswani2017attention} is one solution to solve these problems. The Transformer is based on the self-attention mechanism, has natural advantages for parallelization, and retains relative long-range dependencies. The Transformer model has achieved promising results in MDS tasks \cite{liu2018generating, liu2019hierarchical,li2020leveraging, jin2020multi} and can replace the \textsl{DNN based Model} in Figure \ref{fig:network}. Most of the Transformer based models follow an encoder-decoder structure. Transformer based models can be divided into flat Transformer, hierarchical Transformer, and pre-train language models. 

\vspace{1mm}
\noindent
\textbf{Flat Transformer.} 
Liu et al. \cite{liu2018generating} introduced Transformer to MDS tasks, aiming to generate a Wikipedia article from a given topic and set of references. The authors argue that the encoder-decoder based sequence transduction model cannot cope well with long input documents, so their model selects a series of top-$K$ tokens and feeds them into a Transformer based decoder-only sequence transduction model to generate Wikipedia articles. More specifically, the Transformer decoder-only architecture combines the result from the extractive stage and golden summary into a sentence for training.
To obtain rich semantic representation from different granularity, Jin et al. \cite{jin2020multi} proposed a Transformer based multi-granularity interaction network \textsl{MGSum} and unified extractive and abstractive MDS. Words, sentences and documents are considered as three granular levels of semantic unit connected by a granularity hierarchical relation graph. In the same granularity, a self-attention mechanism is used to capture the semantic relationships. Sentence granularity representation is employed in the extractive summarization, and word granularity representation is adapted to generate an abstractive summary. \textsl{MGSum} employs a fusion gate to integrate and update the semantic representation. Additionally, a spare attention mechanism is used to ensure the summary generator focus on important information. Brazinskas et al. \cite{Brazinskas2020few} created a precedent for few-shot learning for MDS that leverages a Transformer conditional language model and a plug-in network for both extractive and abstractive MDS to overcome rapid overfitting and poor generation problems resulting from naive fine-tuning of large parameter models.

\vspace{1mm}
\noindent 
\textbf{Hierarchical Transformer.}
To handle huge input documents, Yang et al. \cite{liu2019hierarchical} proposed a  two-stage \textsl{Hierarchical Transformer (HT) model} with an inter-paragraph and graph-informed attention mechanism that allows the model to encode multiple input documents hierarchically instead of by simple flat-concatenation.
A logistic regression model is employed to select the top-$K$ paragraphs, which are fed into a local Transformer layer to obtain contextual features. A global Transformer layer mixes the contextual information to model the dependencies of the selected paragraphs. To leverage graph structure to capture cross-document relations, Li et al. \cite{li2020leveraging} proposed an end-to-end Transformer based model \textsl{GraphSum}, based on the HT model. 
In the graph encoding layers, GraphSum extends the self-attention mechanism to the graph-informed self-attention mechanism, which incorporates the graph representation into the Transformer encoding process. Furthermore, the Gaussian function is applied to the graph representation matrix to control the intensity of the graph structure impact on the summarization model. The HT and GraphSum models are both based on the self-attention mechanism leading quadratic memory growth increases with the number of input sequences; to address this issue, Pasunuru et al. \cite{pasunuru2021efficiently} modified the full self-attention with local and global attention mechanism \cite{beltagy2020longformer} to scale the memory linearly. Dual encoders are proposed for encoding truncated concatenated documents and linearized graph information from full documents.

\vspace{1mm}
\noindent 
\textbf{Pre-trained language models (LMs)}. Pre-trained Transformers on large text corpora have shown great successes in downstream NLP tasks including text summarization. 
The pre-trained LMs can be trained on non-summarization or SDS datasets to overcome lack of MDS data \cite{Dzhang2019pegasus, li2020leveraging, pasunuru2021efficiently}.  
Most pre-trained LMs such as BERT \cite{devlin2018bert} and RoBERTa \cite{liu2019roberta} can work well on short sequences. In hierarchical Transformer architecture, replacing the low-level Transformer (token-level) encoding layer with pre-trained LMs helps the model break through length limitations to perceive further information \cite{li2020leveraging}. Inside a hierarchical Transformer architecture, the output vector of the "[CLS]" token can be used as input for high-level Transformer models.
To avoid the self-attention quadratic-memory increment when dealing with document-scale sequences, a Longformer based approach  \cite{beltagy2020longformer}, including local and global attention mechanisms, can be incorporated with pre-trained LMs to scale the memory linearly for MDS \cite{pasunuru2021efficiently}. Another solution for computational issues can be borrowed from SDS is to use a multi-layer Transformer architecture to scale the length of documents allowing pre-trained LMs to encode a small block of text and the information can be shared among the blocks between two successive layers \cite{grail2021globalizing}.
PEGASUS \cite{Dzhang2019pegasus} is a pre-trained Transformer-based encoder-decoder model with gap-sentences generation (GSG) specifically designed for abstractive summarization. GSG shows that masking whole sentences based on importance, instead of through random or lead selection, works well for downstream summarization tasks.

\subsection{Deep Hybrid Models}
Many neural models can be integrated to formalize a more powerful and expressive model. In this section, we summarize the existing deep hybrid models that have proven to be effective for MDS.

\vspace{1mm}
\noindent \textbf{CNN + LSTM + Capsule networks.} 
Cho et al. \cite{cho2019improving} proposed a hybrid model based on the determinantal point processes for semantically measuring sentence similarities. A convolutional layer slides over the pairwise sentences with filters of different sizes to extract low-level features. Capsule networks \cite{sabour2017dynamic,zhao2018investigating} are employed to identify redundant information by transforming the spatial and orientational relationships for high-level representation. The authors also used LSTM to reconstruct pairwise sentences and add reconstruction loss to the final objective function.

\vspace{1mm}
\noindent \textbf{CNN + Bi-LSTM + Multi-layer Perceptron (MLP). } 
Abhishek et al. \cite{singh2018unity} proposed an extractive MDS framework that considers document-dependent and document-independent information. In this model, a CNN with different filters captures phrase-level representation. Full binary trees formed with these salient representation are fed to the recommended Bi-LSTM tree indexer to enable better generalization abilities. A MLP with ReLU function is employed for leaf node transformation. More specifically, the Bi-LSTM tree indexer leverages the time serial power of LSTMs and the compositionality of recursive models to capture both semantic and compositional features. 

\vspace{1mm}
\noindent \textbf{PG networks + Transformer. } 
In generating a summary, it is necessary to consider the information fusion of multiple sentences, especially sentence pairs. Logan et al. \cite{logan2019scoring} found the majority of summary sentences are generated by fusing one or two source sentences; so they proposed a two-stage summarization method that considers the semantic compatibility of sentence pairs. This method joint-scores single sentence and sentence pairs to filter representative from the original documents. Sentences or sentence pairs with high scores are then compressed and rewritten to generate a summary that leverages PG network. This paper uses a Transformer based model to encode both single sentence and sentence pairs indiscriminately to obtain the deep contextual representation of words and sequences.

\begin{table}[]
\caption{Deep Learning based Methods. "Ext", "Abs" and "Hyd" mean extractive, abstractive and hybrid respectively; "FC" and "HC" represent Flat Concatenate, Hierarchical Concatenate respectively. }
\label{tab:Deep_learning_based_methods}
\scalebox{0.8}{
\begin{tabular}{|c|c|c|c|c|c|c|l|l|}
\hline
\multirow{2}{*}{\textbf{Methods}} & \multirow{2}{*}{\textbf{Works}} & \multicolumn{3}{c|}{\begin{tabular}[c]{@{}c@{}} \textbf{Construction} \\ \textbf{Types}\end{tabular}} & \multicolumn{2}{c|}{\begin{tabular}[c]{@{}c@{}}\textbf{Document-level}\\ \textbf{Relationship}\end{tabular}} & \multicolumn{2}{c|}{\begin{tabular}[c]{@{}c@{}}\textbf{Comparison of} \\ \textbf{DL based techniques}\end{tabular}} \\ \cline{3-9}
 &  & Ext & Abs & Hyb & FC & HC & \multicolumn{2}{c|}{Pros and Cons} \\ \hline
\multirow{10}{*}{RNN} & MeanSum \cite{chu2019meansum} &  & \checkmark &  & \checkmark &  & \multicolumn{2}{l|}{\multirow{10}{*}{\begin{tabular}[c]{@{}l@{}} Pros: Can capture sequential \\ relations and syntactic/semantic\\ information from word\\ sequences \\ Cons: Not easy to parallel \\ computing; Highly depending \\ on results from the \\ previous steps \end{tabular}}} \\ \cline{2-7}
 & Zhang et al. \cite{zhang2018adapting} &  & \checkmark &  & \checkmark &  & \multicolumn{2}{l|}{} \\ \cline{2-7}
 & STDS \cite{zheng2019subtopic} & \checkmark &  &  &  & \checkmark & \multicolumn{2}{l|}{} \\ \cline{2-7}
 & ParaFuse\_doc \cite{mir2018abstractive} &  & \checkmark &  &  & \checkmark & \multicolumn{2}{l|}{} \\ \cline{2-7}
 & R2N2 \cite{cao2015ranking} & \checkmark &  &  & \checkmark &  & \multicolumn{2}{l|}{} \\ \cline{2-7}
 & CondaSum \cite{amplayo2019informative} &  &  & \checkmark &  & \checkmark & \multicolumn{2}{l|}{} \\ \cline{2-7}
 & C-Attention \cite{li2017cascaded} &  & \checkmark &  & \checkmark &  & \multicolumn{2}{l|}{} \\ \cline{2-7}
 & Wang et al.\cite{wang2016neural} &  & \checkmark &  & \checkmark &  & \multicolumn{2}{l|}{} \\ \cline{2-7}
 & RL-MMR \cite{mao2020multi} & \checkmark &  &  & \checkmark &  & \multicolumn{2}{l|}{} \\ \cline{2-7}
 & Coavoux et al.\cite{coavoux2019unsupervised} &  & \checkmark &  & \checkmark &  & \multicolumn{2}{l|}{} \\ \hline
\multirow{5}{*}{CNN} & MV-CNN \cite{zhang2016multiview} & \checkmark &  &  & \checkmark &  & \multicolumn{2}{l|}{\multirow{5}{*}{\begin{tabular}[c]{@{}l@{}} Pros: Good parallel computing; \\ Cons: Not good at processing \\ sequential data\end{tabular}}} \\ \cline{2-7}
 & TCSum \cite{cao2017improving} & \checkmark &  &  & \checkmark &  & \multicolumn{2}{l|}{} \\ \cline{2-7}
 & CNNLM \cite{yin2015optimizing} & \checkmark &  &  & \checkmark &  & \multicolumn{2}{l|}{} \\ \cline{2-7}
 & PriorSum \cite{cao2015learning} & \checkmark &  &  & \checkmark &  & \multicolumn{2}{l|}{} \\ \cline{2-7}
 & Angelidis et al.\cite{angelidis2018summarizing} & \checkmark &  &  & \checkmark &  & \multicolumn{2}{l|}{} \\ \hline
\multirow{4}{*}{GNN} & Yasunaga et al.\cite{yasunaga2017graph} & \checkmark &  &  &  & \checkmark & \multicolumn{2}{l|}{\multirow{4}{*}{\begin{tabular}[c]{@{}l@{}} Pros: Can capture cross-document\\ and in-document relations \\ Cons: Inefficient when \\ dealing with large graphs \end{tabular}}} \\ \cline{2-7}
 & SemSentSum \cite{antognini2019learning} & \checkmark &  &  &  & \checkmark & \multicolumn{2}{l|}{} \\ \cline{2-7}
 & Scisummnet \cite{yasunaga2019scisummnet} & \checkmark &  &  &  & \checkmark & \multicolumn{2}{l|}{} \\ \cline{2-7}
 & HDSG \cite{wang2020heterogeneous} & \checkmark &  &  &  & \checkmark & \multicolumn{2}{l|}{} \\ \hline
\multirow{2}{*}{PG} & PG-MMR \cite{logan2018adapting} &  & \checkmark &  & \checkmark &  & \multicolumn{2}{l|}{\multirow{2}{*}{\begin{tabular}[c]{@{}l@{}} Pros: Low redundancy\\ Cons: Hard to train\end{tabular}}} \\ \cline{2-7}
 & Hi-MAP \cite{fabbri-etal-2019-multi} &  & \checkmark &  & \checkmark &  & \multicolumn{2}{l|}{} \\ \hline
\multirow{6}{*}{Transformer} & HT \cite{liu2019hierarchical} &  & \checkmark &  &  & \checkmark & \multicolumn{2}{l|}{\multirow{6}{*}{\begin{tabular}[c]{@{}l@{}} Pros: Good performance; Good\\ parallel computing; Can capture\\ cross-document and \\ in-document relations\\ Cons: Time-consuming; Problems \\ with position encoding \end{tabular}}} \\ \cline{2-7}
 & MGSum \cite{jin2020multi} & \checkmark & \checkmark &  &  & \checkmark & \multicolumn{2}{l|}{} \\ \cline{2-7}
 & FewSum \cite{Brazinskas2020few} & \checkmark & \checkmark &  & \checkmark &  & \multicolumn{2}{l|}{} \\ \cline{2-7}
 & GraphSum \cite{li2020leveraging} &  & \checkmark &  &  & \checkmark & \multicolumn{2}{l|}{} \\ \cline{2-7}
 & Bart-Long \cite{pasunuru2021efficiently} &  & \checkmark &  &  & \checkmark & \multicolumn{2}{l|}{} \\ \cline{2-7}
 & WikiSum \cite{liu2018generating} &  &  & \checkmark & \checkmark &  & \multicolumn{2}{l|}{} \\ \hline
\multirow{3}{*}{Deep Hybid Model} & Cho et al.\cite{cho2019improving} & \checkmark &  &  & \checkmark &  & \multicolumn{2}{l|}{\multirow{3}{*}{\begin{tabular}[c]{@{}l@{}} Pros: Combines the advantages \\ of different DL models\\ Cons: Computationally intensive\end{tabular}}} \\ \cline{2-7}
 & GT-SingPairMix \cite{logan2019scoring} &  & \checkmark &  & \checkmark &  & \multicolumn{2}{l|}{} \\ \cline{2-7}
 & HNet \cite{singh2018unity} & \checkmark &  &  & \checkmark &  & \multicolumn{2}{l|}{} \\ \hline
\end{tabular}}
\end{table}

\subsection{The Variants of Multi-document Summarization}

In this section, we briefly introduce several MDS task variants to give researchers a comprehensive understanding of MDS. These tasks can be modeled as MDS problems and adopt the aforementioned deep learning techniques and neural network architectures.

\vspace{1mm}
\noindent 
\textbf{Query-oriented MDS}
calls for a summary from a set of documents that answers a query. It tries to solve realistic query-oriented scenario problems and only summarizes important information that best answers the query in a logical order \cite{pasunuru2021data}. Specifically, query-oriented MDS combines the information retrieval and MDS techniques. The content that needs to be summarized is based on the given queries.
Liu et al. \cite{liu2019hierarchical} incorporated the query by simply prepending  the query to the top-ranked document during encoding. Pasunuru \cite{pasunuru2021data} involved a query encoder and integrated query embedding into an MDS model, ranking the importance of documents for a given query.

\vspace{1mm}
\noindent  
\textbf{Dialogue summarization}
aims to provide a succinct synopsis from multiple textual utterances of two or more participants, which could help quickly capture relevant information without having to listen to long and convoluted dialogues \cite{liu2019automatic}. Dialogue summary covers several areas, including meetings \cite{zhu2020hierarchical, jia2020how, feng2021dialogue}, email threads \cite{zhang2021emailsum}, medical dialogues  \cite{song2020summarizing, joshi2020dr, enarvi2020generating}, customer service \cite{liu2019automatic} and media interviews \cite{zhu2021mediasum}.  Challenges in dialogue summarization can be summarized into the following seven categories: informal language use, multiple participants, multiple turns, referral and coreference, repetition and interruption, negations and rhetorical questions, role and language change \cite{chen2020multi}.
The flow of the dialogue would be neglected if MDS models are directly applied for dialogue summarization. Liu et al. \cite{liu2019automatic} relied on human annotations to capture the logic of the dialogue. Wu et al. \cite{wu2021controllable} used summary sketch to identify the interaction between speakers and their corresponding textual utterances in each turn. Chen et al. \cite{chen2020multi} proposed a multi-view sequence to sequence based encoder to extract dialogue structure and a multi-view decoder to incorporate different views to generate final summaries.

\vspace{1mm}
\noindent 
\textbf{Stream summarization}
aims to summarize new documents in a continuously growing document stream, such as information from social media. Temporal summarization and real-time summarization (RTS)\footnote{http://trecrts.github.io/} can be seen as a form of stream document summarization. Stream summarization considers both historical dependencies and future uncertainty of the document stream. Yang et al. \cite{yang2020be} used deep reinforcement learning to solve the relevance, redundancy, and timeliness issues in steam summarization. Tan et al. \cite{Tan2017neural} transformed the real time summarization task as a sequential decision making problem and used a LSTM layer and three fully connected neural network layers to maximize the long-term rewards.

\subsection{Discussion} 
In this section, we have reviewed the state-of-the-art works of deep learning based MDS models according to the neural networks applied. Table \ref{tab:Deep_learning_based_methods} summarizes the reviewed works by considering the type of neural networks, construction types, and concatenation methods; and provides a high-level summary of their relative advantages and disadvantages. Transformer based models have been most commonly used in the last three years because they overcome the limitations of CNN's fixed-size receptive field and RNN's inability to parallel process. 
However, deep learning based MDS models face some challenges. Firstly, the complexity of deep learning based models and the data-driven deep learning systems do require more training data, with concomitant increased efforts in data labelling, and computing resources than non-deep learning based methods, which are not time efficient.  
Secondly, deep learning based methods lack linguistic knowledge that can serve as important roles in assisting deep learning based learners to have informative representation and better guide the summary generation. We believe that this is one possible reason that some non-deep learning based MDS methods sometimes show better performance than deep learning based methods \cite{Lu2020multixscience, cao2015learning} as non-deep learning based methods pay more attention to linguistic information. 
We discuss this point in Section \ref{sec: futurework}. Further researches could also be based on techniques adopted in non-deep learning based MDS as reviewed in  \cite{ferreira2014multi, shah2016literature, eswa/El-KassasSRM21}.

\section{Objective Functions}
\label{sec: objective}

In this section, we will take a closer look at different objective functions adopted by various MDS models. In summarization models, objective functions play an important role by guiding the model to achieve specific purposes. To the best of our knowledge, we are the first to provide a comprehensive survey on different objectives of summarization tasks. 

\subsection{Cross-Entropy Objective}

Cross-entropy usually acts as an objective function to measure the distance between two distributions. Many existing MDS models adopt it to measure the difference between the distributions of generated summaries and the golden summaries \cite{cao2015ranking, zhang2016multiview, wang2020heterogeneous, zhang2018adapting, cho2019improving, yasunaga2019scisummnet}. Formally, cross-entropy loss is defined as:

\begin{equation} \small
    L_{CE} = -\sum_{i=1} \mathbf{y_{i}}\log(\mathbf{\hat{y}_{i}}),
\end{equation}

\noindent where $\mathbf{y_i}$ is the target score from golden summaries and machine-generated summaries, and $\mathbf{\hat{y}_{i}}$ is the predicted estimation from the deep learning based models. Different from calculations in other tasks, such as text classification, in summarization tasks, $\mathbf{y_i}$ and $\mathbf{\hat{y}_{i}}$ have several methods to calculate. $\mathbf{\hat{y}_{i}}$ usually is calculated by Recall-Oriented Understudy for Gisting Evaluation (ROUGE) (Please refer to Section \ref{sec: evaluation}). For example, ROUGE-1\cite{antognini2019learning}, ROUGE-2 \cite{liu2019hierarchical} or the normalized average of ROUGE-1 and ROUGE-2 scores \cite{yasunaga2017graph} could be adopted to compute the ground truth score between the selected sentences and golden summary.

\subsection{Reconstructive Objective}

Reconstructive objectives are used to train a distinctive representation learner by reconstructing the input vectors in an unsupervised learning manner. The objective function is defined as:

\begin{equation} \small
    L_{Rec} = \left \|  \mathbf{x_{i}}-\phi^{\prime}(\phi(\mathbf{x_{i}};\theta);\theta^{\prime})\right \|_{*} ,
\end{equation}

\vspace{1mm}
\noindent where $\mathbf{x_{i}}$ represents the input vector; $\phi$ and $\phi^{\prime}$ represent the encoder and decoder with $\theta$ and $\theta^{\prime}$ as their parameters respectively, $||\cdot||_*$ represents norm (* stands for 0, 1, 2, ..., infinity). $L_{Rec}$ is a measuring function to calculate the distance between source documents and their reconstructive outputs.
Chu et al. \cite{chu2019meansum} used a reconstructive loss to constrain the generated text into the natural language domain, reconstructing reviews in a token-by-token manner. Moreover, this paper also proposes a variant termed \textit{reconstruction cycle loss}. By using the variant, the reviews are encoded into a latent space to further generate the summary, and the summary is then decoded to the reconstructed reviews to form another reconstructive closed-loop.
An unsupervised learning loss was designed by Li et al. \cite{li2017cascaded} to reconstruct the condensed output vectors to the original input sentence vectors with $L_{2}$ distance. This paper further constrains the condensed output vector with a $L_{1}$ regularizer to ensure sparsity. Similarly, Zheng et al. \cite{zheng2019subtopic} adopted a bi-directional GRU encoder-decoder framework to reconstruct both news sentences and comment sentences in a word sequence manner. 
Liu et al. \cite{liu2018generating} used reconstruction within the abstractive stage of a two-stage strategy to alleviate the problem introduced by long input documents. Both input and output sequences are concatenated to predict the next token to train the abstractive model. There are also some variants, such as leveraging the latent vectors of variational auto-encoder for reconstruction to capture better representation. Li et al. \cite{li2017reader} introduced three individual reconstructive losses to consider both news reconstruction and comments reconstruction separately, along with a variational auto-encoder lower bound. Bravzinskas et al. \cite{bravzinskas2019unsupervised} utilized a variational auto-encoder to generate the latent vectors of given reviews, where each review is reconstructed by the latent vectors combined with other reviews.

\subsection{Redundancy Objective}

Redundancy is an important objective to minimize the overlap between semantic units in a machine-generated summary. By using this objective, models are encouraged to maximize information coverage. Formally,
\vspace{1mm}
\begin{equation} \small
    L_{Red} = Sim(\mathbf{x_{i}}, \mathbf{x_{j}}),
\end{equation}

\vspace{1mm}
\noindent where $Sim(\cdot)$ is the similarity function to measure the overlap between different $\mathbf{x_{i}}$ and $\mathbf{x_{j}}$, which can be phrases, sentences, topics or documents. The redundancy objective is often treated as an auxiliary objective combined with other loss functions. 
Li et al. \cite{li2017cascaded} penalized phrase pairs with similar meanings to eliminate the redundancy. Nayeem et al. \cite{mir2018abstractive} used the redundancy objective to avoid generating repetitive phrases, constraining a sentence to appear only once while maximizing the scores of important phrases. Zheng et al. \cite{zheng2019subtopic} adopted a redundancy loss function to measure overlaps between subtopics; intuitively, smaller overlaps between subtopics resulted in less redundancy in the output domain. 
Yin et al. \cite{yin2015optimizing} proposed a redundancy objective to estimate the diversity between different sentences.

\subsection{Max Margin Objective}

Max Margin Objectives (MMO) are also used to empower the MDS models to learn better representation. The objective function is formalized as:
\vspace{1mm}
\begin{equation} \small
    L_{Margin} =\max \left(0, f(\mathbf{x_{i}};\theta)-f(\mathbf{x_{j}};\theta)+\gamma\right),
\end{equation}

\vspace{1mm}
\noindent where $\mathbf{x_{i}}$ and $\mathbf{x_{j}}$ represent the input vectors, $\theta$ are parameters of the model function $f(\cdot)$, and $\gamma$ is the margin threshold. The MMO aims to force function $f(\mathbf{x_{i}};\theta)$ and function $f(\mathbf{x_{j}};\theta)$ to be separated by a predefined margin $\gamma$. In Cao et al. \cite{cao2017improving}, a MMO is designed to constrain a pair of randomly sampled sentences with different salience scores -- the one with higher score should be larger than the other one more than a marginal threshold. Two max margin losses are proposed in Zhong et al. \cite{zhong2020extractive}: a margin-based triplet loss that encouraged the model to pull the golden summaries semantically closer to the original documents than to the machine-generated summaries; and a pair-wise margin loss based on a greater margin between paired candidates with more disparate ROUGE score rankings.

\subsection{Multi-Task Objective}

Supervision signals from MDS objectives may not be strong enough for representation learners, so some works seek other supervision signals from multiple tasks. A general form is as follows:
\vspace{1mm}
\begin{equation} \small
    L_{Mul} = L_{Summ} + L_{Other}, 
\end{equation}

\vspace{1mm}
\noindent where $L_{Summ}$ is the loss function of MDS tasks, and $L_{Other}$ is the loss function of an auxiliary task. Angelidis et al. \cite{angelidis2018summarizing} assumed that the aspect-relevant words not only provides a reasonable basis for model aspect reconstruction, but also a good indicator for product domain. 
Similarly, multi-task classification was introduced by Cao et al. \cite{cao2017improving}. Two models are maintained: text classification and text summarization models. In the first model, CNN is used to classify text categories and cross-entropy loss is used as the objective function. The summarization model and the text classification model share parameters and pooling operations, so are equivalent to the shared document vector representation. Coavoux et al. \cite{coavoux2019unsupervised} jointly optimized the model from a language modeling objective and two other multi-task supervised classification losses, which are polarity loss and aspect loss.

\subsection{Other Types of Objectives}
There are many other types of objectives in addition to those mentioned above. Cao et al. \cite{cao2015learning} proposed using ROUGE-2 to calculate the sentence saliency scores and the model tries to estimate this saliency with linear regression. 
Yin et al. \cite{yin2015optimizing} suggested summing the squares of the prestige vectors calculated by the PageRank algorithm to identify sentence importance. 
Zhang et al. \cite{zhang2016multiview} proposed an objective function by ensembling individual scores from multiple CNN models; besides the cross-entropy loss, a consensus objective is adopted to minimize disagreement between each pair of classifiers. 
Amplay et al. \cite{amplayo2019informative} used two objectives in the abstract module: the first to optimize the generation probability distribution by maximizing the likelihood; and the second to constrain the model output to be close to its golden summary in the encoding space, as well as being distant from the random sampled negative summaries. 
Chu et al. \cite{chu2019meansum} designed a similarity objective that shares the encoder and decoder weights within the auto-encoder module, while in the summarization module, the average cosine distance indicates the similarity between the generated summary and the reviews. A variant similarity objective termed \textit{early cosine objective} is further proposed to compute the similarity in a latent space which is the average of the cells states and hidden states to constrain the generated summaries semantically close to reviews.

\subsection{Discussion} 
In MDS, cross-entropy is the most commonly adopted objective function that bridges the predicted candidate summaries and the golden summaries by treating the golden summaries as strong supervision signals. However, adopting cross-entropy loss alone may not lead the model to achieve good performance since the supervisory signal for cross-entropy objective is not strong enough by itself to effectively learn good representation. Several other objectives can thus serve as complements, e.g., reconstruction objectives offer a view from the unsupervised learning perspective; the redundancy objective constrains models from generating redundant content; while max-margin objectives require a step-change improvements from previous versions. By using multiple objectives, model optimization could be conducted with the input documents themselves if the manual annotation is scarce. The models that  adopt multi-task objectives explicitly define multiple auxiliary tasks to assist the main summarization task for better generalization, and provide various constraints from different angles that lead to better model optimization.

\section{Evaluation metrics}
\label{sec: evaluation}
Evaluation metrics are used to measure the effectiveness of a given method objectively, so well-defined evaluation metrics are crucial to MDS research. We classify the existing evaluation metrics in two categories and will discuss each category in detail: (1) ROUGE: the most commonly used evaluation metrics in the summarization community; and (2) other evaluation metrics that have not been widely used in MDS research to date.

\subsection{ROUGE} 
\textsl{Recall-Oriented Understudy for Gisting Evaluation (ROUGE)} \cite{lin2004rouge} is a collection of evaluation indicators that is one of the most essential metrics for many natural language processing tasks, including machine translation and text summarization. ROUGE obtains prediction/ground-truth similarity scores through comparing automatically generated summaries with a set of corresponding human-written references. ROUGE has many variants to measure candidate abstracts in a variety of ways \cite{lin2004rouge}. The most commonly used ones are ROUGE-N and ROUGE-L.

\vspace{1mm}
\noindent \textbf{ROUGE-N} (\textit{ROUGE with n-gram co-occurrence statistics} ) measures a n-gram recall between reference summaries and their corresponding candidate summaries \cite{lin2004rouge}. Formally, ROUGE-N can be calculated as:
\vspace{1mm}
\begin{equation} \small
    ROUGE\text{-}N=\frac{ \sum_{Sum \in \{Ref\}} \sum_{gram_n \in Sum} Count_{match}(gram_n) } { \sum_{Sum \in \{Ref\}} \sum_{gram_n \in Sum} Count(gram_n) },
\end{equation}

\vspace{1mm}
\noindent where $Ref$ and $Sum$ are reference summary and machine-generated summary, $n$ represents the length of n-gram, and $Count_{match}(gram_n)$ represents the maximum number of n-grams in the reference summary and corresponding candidates. The numerator of ROUGE-N is the number of n-grams owned by both the reference and generated summary, while the denominator is the total number of n-grams occurring in the golden summary. The denominator could also be set to the number of candidate summary n-grams to measure precision; however, ROUGE-N mainly focuses on quantifying recall, so precision is not usually calculated. ROUGE-1 and ROUGE-2 are special cases of ROUGE-N that are usually chosen as best practices and represent the unigram and bigram, respectively.

\vspace{1mm}
\noindent \textbf{ROUGE-L}
(\textsl{ROUGE with Longest Common Subsequence}) adopts the longest common subsequence algorithm to count the longest matching vocabularies \cite{lin2004rouge}. Formally, ROUGE-L is calculated using:

\begin{equation} \small
\label{equ:rouge-l3}
    F_{lcs}=\frac{(1+\beta^2)R_{lcs}P_{lcs}} {R_{lcs} + \beta^2P_{lcs}},
\end{equation}
where 
\begin{equation} \small
\label{equ:rouge-l1}
    R_{lcs}=\frac{LCS(Ref, Sum)} {m},
\end{equation}
and
\begin{equation} \small
\label{equ:rouge-l2}
    P_{lcs}=\frac{LCS(Ref, Sum)} {n}.
\end{equation}

\vspace{1mm}
\noindent where LCS($\cdot$) represents the longest common subsequence function. ROUGE-L is termed as LCS-based F-measure as it is obtained from LCS-Precision $P_{lcs}$ and LCS-Recall $R_{lcs}$. $\beta$ is the balance factor between $R_{lcs}$ and $P_{lcs}$. It can be set by the fraction of $P_{lcs}$ and $R_{lcs}$; by setting $\beta$ to a big number, only $R_{lcs}$ is considered. The use of ROUGE-L enables measurement of the similarity of two text sequences at sentence-level. ROUGE-L also has the advantage of automatically deciding the n-gram without extra manual input, since the calculation of LCS empowers the model to count grams adaptively.

\vspace{1mm}
\noindent \textbf{Other ROUGE Based Metrics.}
\textsl{ROUGE-W} \cite{lin2004rouge} is proposed to weight consecutive matches to better measure semantic similarities between two texts. \textsl{ROUGE-S} \cite{lin2004rouge} stands for ROUGE with Skip-bigram co-occurrence statistics that allows the bigram to skip arbitrary words. An extension of ROUGE-S, \textsl{ROUGE-SU} \cite{lin2004rouge} refers to ROUGE with Skip-bigram plus Unigram-based co-occurrence statistics and is able to be obtained from ROUGE-S by adding a begin-of-sentence token at the start of both references and candidates. \textsl{ROUGE-WE} \cite{ng2015better} is proposed to further extend ROUGE by measuring the pair-wise summary distances in word embeddings space. In recent years, more ROUGE-based evaluation models have been proposed to compare golden and machine-generated summaries, not just according to their literal similarity, but also considering semantic similarity \cite{shafieibavani2018graph, zhao2019moverscore, zhang2020bertscore}. In terms of the ROUGE metric for multiple golden summaries, the Jackknifing procedure (similar to K-fold validation) has been introduced \cite{lin2004rouge}. The $M$ best scores are computed from sets composed of $M$-1 reference summaries and the final ROUGE-N is the average of $M$ scores. This procedure can also be applied to ROUGE-L, ROUGE-W and ROUGE-S.

\subsection{Other Evaluation Metrics}
Besides \textsl{ROUGE}-based \cite{lin2004rouge} metrics, other evaluation metrics for MDS exist, but have received less attention than ROUGE. We hope this section will give researchers and practitioners a holistic view of alternative evaluation metrics in this field. Based on the mode of summaries matching, we divide the evaluation metrics into two groups: lexical matching metrics and semantic matching metrics.

\begin{table}[]
\caption{Advantages and disadvantages of different evaluation metrics.}\label{tab:evaluations}
\scalebox{0.8}{
\begin{tabular}{|cc|l|l|}
\hline
\multicolumn{2}{|c|}{\textbf{Evaluation Metrics}} & \multicolumn{1}{c|}{\textbf{Advantages}} & \multicolumn{1}{c|}{\textbf{Disadvantages}} \\ \hline
\multicolumn{1}{|c|}{\multirow{6}{*}{\begin{tabular}[c]{@{}c@{}} \\ \\ \\ \\ \\ Lexical \\ Matching \\ Metrics\end{tabular}}} & ROUGE & \begin{tabular}[c]{@{}l@{}} \tabitem Widely used \\ \tabitem Intuitive \\ \tabitem Easily computed\end{tabular} & \begin{tabular}[c]{@{}l@{}} \tabitem Cannot measure texts \\ \ \ semantically \\ \tabitem Exact matching\end{tabular} \\ \cline{2-4} 
\multicolumn{1}{|c|}{} & BLEU & \begin{tabular}[c]{@{}l@{}} \tabitem Intuitive \\ \tabitem Easily computed  \\ \tabitem High correlations with \\ \ \ human judgments\end{tabular} & \begin{tabular}[c]{@{}l@{}} \tabitem Cannot measure texts \\ \ \ semantically \\ \tabitem Cannot deal with \\ \ \ languages lacking word \\ \ \ boundaries\end{tabular} \\ \cline{2-4} 
\multicolumn{1}{|c|}{} & Perplexity & \begin{tabular}[c]{@{}l@{}} \tabitem Easily computed \\ \tabitem Intuitive \end{tabular} & \begin{tabular}[c]{@{}l@{}} \tabitem Sensitive to certain \\ \ \ symbols and words\end{tabular} \\ \cline{2-4} 
\multicolumn{1}{|c|}{} & Pyramid & \begin{tabular}[c]{@{}l@{}} \tabitem High correlations with \\ \ \ human judgments\end{tabular} & \begin{tabular}[c]{@{}l@{}} \tabitem Requires manually \\ \ \ extraction of units \\ \tabitem Bias results easily\end{tabular} \\ \cline{2-4} 
\multicolumn{1}{|c|}{} & Responsiveness & \begin{tabular}[c]{@{}l@{}} \tabitem Consider both content and \\ \ \ linguistic quality \\ \tabitem Can be calculated without \\ \ \ reference \end{tabular} & \tabitem Not widely adopted \\ \cline{2-4} 
\multicolumn{1}{|c|}{} & \begin{tabular}[c]{@{}c@{}}Data \\ Statistics\end{tabular} & \begin{tabular}[c]{@{}l@{}} \tabitem Can measure the density \\ \ \ and coverage of summary\end{tabular} & \begin{tabular}[c]{@{}l@{}} \tabitem Cannot measure texts \\ \ \ semantically\end{tabular} \\ \hline
\multicolumn{1}{|c|}{\multirow{7}{*}{\begin{tabular}[c]{@{}c@{}} \\ \\ \\ \\ \\ \\ Semantic \\ Matching  \\ Metrics\end{tabular}}} & MEREOR & \tabitem Consider non-exact matching & \begin{tabular}[c]{@{}l@{}} \tabitem Sensitive to length \end{tabular} \\ \cline{2-4} 
\multicolumn{1}{|c|}{} & SUPERT & \begin{tabular}[c]{@{}l@{}} \tabitem Can measuring texts semantic \\ \ \ similarity\end{tabular} & \tabitem Not widely adopted \\ \cline{2-4} 
\multicolumn{1}{|c|}{} & \begin{tabular}[c]{@{}c@{}}Preferences\\ based Metric\end{tabular} & \begin{tabular}[c]{@{}l@{}} \tabitem Does not depend on the \\ \ \ golden summaries\end{tabular} & \begin{tabular}[c]{@{}l@{}}\tabitem Require human \\ \ \ annotations\end{tabular} \\ \cline{2-4} 
\multicolumn{1}{|c|}{} & BERTScore & \begin{tabular}[c]{@{}l@{}} \tabitem Semantically measure texts to \\ \ \ some extent \\ \tabitem Mimic human evaluation\end{tabular} & \begin{tabular}[c]{@{}l@{}} \tabitem High computational \\ \ \ demands\end{tabular} \\ \cline{2-4} 
\multicolumn{1}{|c|}{} & MoverScore & \begin{tabular}[c]{@{}l@{}} \tabitem Semantically measure texts to \\ \ \ some extent \\ \tabitem More similar to human \\ \ \ evaluation by adopting earth \\ \ \ mover’s distance\end{tabular} & \begin{tabular}[c]{@{}l@{}} \tabitem High computational \\ \ \ demands\end{tabular} \\ \cline{2-4} 
\multicolumn{1}{|c|}{} & Importance & \begin{tabular}[c]{@{}l@{}} \tabitem Combining redundancy, \\ \ \ relevance and informativeness \\ \tabitem Theoretically supported\end{tabular} & \begin{tabular}[c]{@{}l@{}} \tabitem Non-trivial for \\ \ \ implementation\end{tabular} \\ \cline{2-4} 
\multicolumn{1}{|c|}{} & \begin{tabular}[c]{@{}c@{}}Human \\ Evaluation\end{tabular} & \begin{tabular}[c]{@{}l@{}} \tabitem Can accurately and \\ \ \ semantically measure texts\end{tabular} & \begin{tabular}[c]{@{}l@{}}\tabitem Require human \\ \ \ annotations\end{tabular} \\ \hline
\end{tabular}}
\end{table}

\vspace{1mm}
\noindent \textbf{Lexical Matching Metrics.}
\textsl{BLEU} \cite{papineni2002bleu} is a commonly used vocabulary-based evaluation metric that provides a precision-based evaluation indicator, as opposed to ROUGE that mainly focuses on recall. 
\textsl{Perplexity} \cite{jelinek1977perplexity} is used to evaluate the quality of the language model by calculating the negative log probability of a word's appearance. A low perplexity on a test dataset is a strong indicator of a summary's high grammatical quality because it measures the probability of words appearing in sequences.
Based on \textsl{Pyramid} \cite{nenkova2007pyramid} calculation, the abstract sentences are manually divided into several Summarization Content Units (SCUs), each representing a core concept formed from a single word or phrase/sentence. After sorting SCUs in order of importance to form the \textsl {Pyramid}, the quality of automatic summarization is evaluated by calculating the number and importance of SCUs included in the document \cite{nenkova2004evaluating}. Intuitively, more important SCUs exist at higher levels of the pyramid. Although \textsl{Pyramid} shows strong correlation with human judgment, it requires professional annotations to match and evaluate SCUs in generated and golden summaries. Some recent works focus on the construction of \textsl{Pyramid} \cite{passonneau2013automated, yang2016peak, hirao2018automatic, gao2019automated, shapira2019crowdsourcing}.
\textsl{Responsiveness} \cite{louis2013automatically} measures content selection and linguistic quality of summaries by directly rating scores.
Additionally, the assessments are calculated without reference to model summaries. 
\textsl{Data Statistics} \cite{Max2018Newroom} contain three evaluation metrics: extractive fragment coverage measures the novelty of generated summaries by calculating the percentage of words in the summary that are also present in source documents; extractive fragment density measures the average length of the extractive block to which each word in the summary belongs; and compression ratio compares the word numbers in the source documents and generated summary.

\vspace{1mm}
\noindent \textbf{Semantic Matching Metrics.}
\textsl{METEOR} (Metric for Evaluation of Translation with Explicit Ordering) \cite{banerjee2005meteor} is an improvement to BLEU. The main idea behind METEOR is that while candidate summaries can be correct with similar meanings, they are not exactly matched with references. In such a case, WordNet\footnote{https://wordnet.princeton.edu/} is introduced to expand the synonym set, and the word form is also taken into account.
\textsl{SUPERT} \cite{gao2020supert} is an unsupervised MDS evaluation metric that measures the semantic similarity between the pseudo-reference summary and the machine-generated summary. \textsl{SUPERT} obviates the need for human annotations by not referring to golden summaries. Contextualized embeddings and soft token alignment techniques are leveraged to select salient information from the input documents to evaluate summary quality.
\textsl{Preferences based Metric} \cite{zopf2018estimating} is a pairwise sentence preference-based evaluation model and it does not depend on the golden summaries. The underlying premise is to ask annotators about their pair-wise preferences rather than writing complex golden summaries, and are much easier and faster to obtain than traditional reference summary-based evaluation models.
\textsl{BERTScore} \cite{zhang2020bertscore} computes a similarity score for each token within the candidate sentence and the reference sentence. It measures the soft overlap of two texts' BERT embeddings.
\textsl{MoverScore} \cite{zhao2019moverscore} adopts a distance to evaluate the agreement between two texts in the context of BERT and ELMo word embeddings. This proposed metric has a high correlation with human judgment of text quality by adopting earth mover's distance.
\textsl{Importance} \cite{peyrard2019simple} is a simple but rigorous evaluation metric from the aspect of information theory. It is a final indicator calculated from the three aspects: \textsl{Redundancy}, \textsl{Relevance}, and \textsl{Informativeness}. A good summary should have low \textit{Redundancy} and high \textit{Relevance} and high \textit{Informativeness}. 
The cluster of \textsl{Human Evaluation} is used to supplement automatic evaluation on relatively small instances. Annotators evaluate the quality of machine-generated summaries by rating \textsl {Informativeness, Fluency, Conciseness, Readability, Relevance.} Model ratings are usually computed by averaging the rating on all selected summary pairs.

\subsection{Discussion} We summarize the advantages and disadvantages of above-mentioned evaluation metrics in Table \ref{tab:evaluations}. Although there are many evaluation metrics for MDS, the indicators of the ROUGE series are generally accepted by the summarization community. Almost all the research works utilize ROUGE for evaluation, while other evaluation indicators are just for assistance currently. Among the ROUGE family, ROUGE-1, ROUGE-2 and ROUGE-L are the most commonly used evaluation metrics. In addition, there are plenty of existing evaluation metrics in other natural language processing tasks that could be potentially adjusted for MDS tasks, such as  \textsl{efficiency}, \textsl{effectiveness} and \textsl{coverage} from information retrieval.

\section{Datasets}\label{sec: datasets} 

Compared to SDS tasks, large-scale MDS datasets, which contain more general scenarios with many downstream tasks, are relatively scarce. In this section, we present our investigation on the 10 most representative datasets commonly used for MDS and its variant tasks.

\vspace{1mm}
\noindent \textbf{DUC \& TAC.}
DUC\footnote{http://duc.nist.gov/} (Document Understanding Conference) provides official text summarization competitions each year from 2001-2007 to promote summarization research. DUC changed its name to Text Analysis Conference (TAC)\footnote{ http://www.nist.gov/tac/} in 2008. Here, the DUC datasets refer to the data collected from 2001-2007; the TAC datasets refer to the dataset after 2008. Both DUC and TAC datasets are from the news domains, including various topics such as politics, natural disaster and biography. Nevertheless, as shown in Table \ref{tab:datasets}, the DUC and TAC datasets provide small datasets for model evaluation that only include hundreds of news documents and human-annotated summaries. Of note, the first sentence in a news item is usually information-rich that renders bias in the news datasets, so it fails to reflect the structure of natural documents in daily lives. These two datasets are on a relatively small scale and not ideal for large-scale deep neural based MDS model training and evaluation.

\begin{table}[]
\caption{Comparison of Different Datasets. In the table, ``Ave'', ``Summ'', ``Len'', ``bus'',``rev'' and ``\#''  represent average, summary, length, business, reviews and numbers respectively; ``Docs'' and ``sents'' mean documents and sentences respectively.
}\label{tab:datasets}
\scalebox{0.8}{
\begin{tabular}{|c||c|c||c|c||c|}
\hline
\textbf{Datasets}                                                     & \textbf{Cluster \#} & \textbf{Document \#}                                                                                                                              & \textbf{Summ \#}       & \textbf{Ave Summ Len}                                                                                                        & \textbf{Topic}                                                       \\ \hline
DUC01                                                        & 30         & 309 docs                                                                                                                                 & 60 summ          & 100 words                                                                                                             & News                                                        \\ \hline
DUC02                                                        & 59         & 567 docs                                                                                                                                 & 116 summ         & 100 words                                                                                                             & News                                                        \\ \hline
DUC03                                                        & 30         & 298 docs                                                                                                                                 & 120 summ         & 100 words                                                                                                             & News                                                        \\ \hline
DUC04                                                        & 50         & 10 docs / cluster                                                                                                                        & 200 summ         & 665 bytes                                                                                                             & News                                                        \\ \hline
DUC05                                                        & 50         & 25-50 docs / cluster                                                                                                                     & 140 summ         & 250 words                                                                                                             & News                                                        \\ \hline
DUC06                                                        & 50         & 25 docs / cluster                                                                                                                        & 4 summ / cluster & 250 words                                                                                                             & News                                                        \\ \hline
DUC07                                                        & 45         & 25 docs / cluster                                                                                                                        & 4 summ / cluster & 250 words                                                                                                             & News                                                        \\ \hline
TAC 2008                                                     & 48         & 10 docs / cluster                                                                                                                        & 4 summ / cluster & 100 words                                                                                                             & News                                                        \\ \hline
TAC 2009                                                     & 44         & 10 docs / cluster                                                                                                                        & 4 summ / cluster & 100 words                                                                                                             & News                                                        \\ \hline
TAC 2010                                                     & 46         & 10 docs / cluster                                                                                                                        & 4 summ / cluster & 100 words                                                                                                             & News                                                        \\ \hline
TAC 2011                                                     & 44         & 10 docs / cluster                                                                                                                        & 4 summ / cluster & 100 words                                                                                                             & News                                                        \\ \hline
OPOSUM                                                       & 60         & 600 rev                                                                                                                              & 1 summ / cluster & 100 words                                                                                                             & \begin{tabular}[c]{@{}c@{}}Amazon\\    reviews\end{tabular} \\ \hline
WikiSum                                                      & -          & \begin{tabular}[c]{@{}c@{}}train / val / test\\    1579360 / 38144 / 38205\end{tabular}                                                  & 1 summ / cluster & 139.4 tokens/ summ                                                                                                 & Wikipedia                                                   \\ \hline
Multi-News                                                   & -          & \begin{tabular}[c]{@{}c@{}}train / val / test\\    44972 / 5622 / 5622\\    2-10 docs / cluster\end{tabular}                 & 1 summ / cluster & \begin{tabular}[c]{@{}c@{}}263.66 words / summ\\    9.97 sents / summ\\    262 tokens / summ\end{tabular} & News                                                        \\ \hline
Opinosis                                                     & 51         & 6457 rev                                                                                                                            & 5 summ / cluster & -                                                                                                                     & \begin{tabular}[c]{@{}c@{}}Site\\    reviews\end{tabular}   \\ \hline
\begin{tabular}[c]{@{}c@{}}Rotten\\    Tomatoes\end{tabular} & 3731       & 99.8 rev / cluster                                                                                                                   & 1 summ / cluster & 19.6 tokens / summ                                                                                                 & \begin{tabular}[c]{@{}c@{}}Movie\\    reviews\end{tabular}  \\ \hline
Yelp                                                         & -          & \begin{tabular}[c]{@{}c@{}}train / val / test\\    bus: 10695 / 1337 / 1337\\    rev: 1038184 / 129856 / 129840\end{tabular} & -                & -                                                                                                                     & \begin{tabular}[c]{@{}c@{}}Customer\\    reviews\end{tabular}   \\ \hline
Scisumm                                                      & 1000       & \begin{tabular}[c]{@{}c@{}}21 - 928 cites / paper\\    15 sents / refer\end{tabular}                                               & 1 summ / cluster & 151 words                                                                                                         & \begin{tabular}[c]{@{}c@{}}Science\\    Paper\end{tabular}  \\ \hline
WCEP                                                         & 10200      & 235 docs / cluster                                                                                                                   & 1 summ / cluster & 32 words                                                                                                          & Wikipedia                                                   \\ \hline
Multi-XScience                                                   & -          & \begin{tabular}[c]{@{}c@{}}train / val / test\\    30369 / 5066 / 5093 \end{tabular}                 & 1 summ / cluster & \begin{tabular}[c]{@{}c@{}} 116.44 words / summ\end{tabular} & \begin{tabular}[c]{@{}c@{}}Science\\    Paper\end{tabular}  \\ \hline
\end{tabular}
}
\end{table}

\vspace{1mm}
\noindent \textbf{OPOSUM.}
OPOSUM \cite{angelidis2018summarizing} collects multiple reviews of six product domains from Amazon. This dataset not only contains multiple reviews and corresponding summaries but also products' domain and polarity information. The latter information could be used as the auxiliary supervision signals.

\vspace{1mm}
\noindent \textbf{WikiSum.}
WikiSum \cite{liu2018generating} targets abstractive MDS. For a specific Wikipedia theme, the documents cited in Wikipedia articles or the top-10 Google search results (using the Wikipedia theme as query) are seen as the source documents. Golden summaries are the real Wikipedia articles. However, some of the URLs are not available and can be identical to each other in parts. To remedy these problems, Liu et al. \cite{liu2019hierarchical} cleaned the dataset and deleted duplicated examples, so here we report statistical results from \cite{liu2019hierarchical}.

\vspace{1mm}
\noindent \textbf{Multi-News.}
Multi-News \cite{fabbri-etal-2019-multi} is a relatively large-scale dataset in the news domain; the articles and human-written summaries are all from the Web\footnote{http://newser.com}. This dataset includes 56,216 article-summary pairs and contain trace-back links to the original documents. Moreover, the authors compared the Multi-News dataset with prior datasets in terms of coverage, density, and compression, revealing that this dataset has various arrangement styles of sequences.

\vspace{1mm}
\noindent \textbf{Opinosis.}
The Opinosis dataset \cite{ganesan2010opinosis} contains reviews of 51 topic clusters collected from TripAdvisor\footnote{https://www.tripadvisor.com/}, Amazon\footnote{https://www.amazon.com.au/}, and Edmunds\footnote{https://www.edmunds.com/}. For each topic, approximately 100 sentences on average are provided and the reviews are fetched from different sources. For each cluster, five professional written golden summaries are provided for the model training and evaluation.

\vspace{1mm}
\noindent \textbf{Rotten Tomatoes.}
The Rotten Tomatoes dataset \cite{wang2016neural} consists of the collected reviews of 3,731 movies from the Rotten Tomato website\footnote{http://rottentomatoes.com}. The reviews contain both professional critics and user comments. For each movie, a one-sentence summary is created by professional editors.

\vspace{1mm}
\noindent \textbf{Yelp.}
Chu et al. \cite{chu2019meansum} proposed a dataset named Yelp based on the Yelp Dataset Challenge. This dataset includes multiple customer reviews with five-star ratings. The authors provided 100 manual-written summaries for model evaluation using Amazon Mechanical Turk (AMT), within which every eight input reviews are summarized into one golden summary.

\vspace{1mm}
\noindent \textbf{Scisumm.}
Scisumm dataset \cite{yasunaga2019scisummnet} is a large, manually annotated corpus for scientific document summarization. The input documents are a scientific publication, called the reference paper, and multiple sentences from the literature that cite this reference paper. In the SciSumm dataset, the 1,000 most cited papers from the ACL Anthology Network \cite{radev2013acl} are treated as reference papers, and an average 15 citation sentences are provided after cleaning. For each cluster, one golden summary is created by five NLP-based Ph.D. students or equivalent professionals.

\vspace{1mm}
\noindent \textbf{WCEP.}
The Wikipedia Current Events Portal dataset (WCEP) \cite{demian2020largescale} contains human-written summaries of recent news events. Similar articles are provided by searching similar articles from Common Crawl News dataset\footnote{https://commoncrawl.org/2016/10/news-dataset-available/} to extend the inputs to obtain large-scale news articles. Overall, the WCEP dataset has good alignment with the real-world industrial use cases.

\vspace{1mm}
\noindent 
\textbf{Multi-XScience.} The source data of Multi-XScience \cite{Lu2020multixscience} are from Arxiv and Microsoft academic graphs and this dataset is suitable for abstractive MDS. Multi-XScience contains fewer positional and extractive biases than WikiSum and Multi-News datasets, so the drawback of obtaining higher scores from a copy sentence at a certain position can be partially avoided.

\vspace{1mm}
\noindent \textbf{Datasets for MDS Variants.}
The representative query-oriented MDS datasets are Debatepedia \cite{nema2017diversity}, AQUAMUSE \cite{kulkarni2020aquamuse}, and QBSUM \cite{zhao2021qbsum}. The representative dialogue summarization datasets are DIALOGSUM \cite{chen2021dialogsumm}, AMI \cite{carletta2005the}, MEDIASUM \cite{zhu2021mediasum}, and QMSum \cite{zhong2021qmsum}. RTS is a track at the Text Retrieval Conference (TREC) which provides several RTS datasets\footnote{http://trecrts.github.io/}. Tweet Contextualization track \cite{Patrice2016INEX} (2012-2014) is derived from the INEX 2011 Question Answering Track, that focuses on more NLP-oriented tasks and moves to MDS.

\vspace{2mm}
\noindent \textbf{Discussion.} 
Table \ref{tab:datasets} compares 20 MDS datasets based on the numbers of clusters and documents; the number and the average length of summaries; and the field to which the dataset belongs. Currently, the main areas covered by the MDS datasets are news (60$\%$), scientific papers (10$\%$) and Wikipedia (10$\%$). In early development of the MDS tasks, most studies were performed on the DUC and TAC datasets. However, the size of these datasets is relatively small, and thus not highly suitable for training deep neural network models. Datasets on news articles are also common, but the structure of news articles (highly compressed information in the first paragraph or first sentence of each paragraph) can cause positional and extractive biases during training. In recent years, large-scale datasets such as WikiSum and Multi-News datasets have been developed and used by researchers to meet training requirements, reflecting the rising trend of data-driven approaches.

\vspace{1mm}
\section{Future research directions and open issues}
\label{sec: futurework}

Although existing works have established a solid foundation for MDS it is a relatively understudied field compared with SDS and other NLP topics.
Summarizing on multi-modal data, medical records, codes, project activities and MDS combining with Internet of Things \cite{wei2020the} have still received less attention. 
Actually, MDS techniques are beneficial for a variety of practical applications, including generating Wikipedia articles, summarizing news, scientific papers, and product reviews, and individuals, industries have a huge demand for compressing multiple related documents into high-quality summaries. This section outlines several prospective research directions and open issues that we believe are critical to resolve in order to advance the field.

\subsection{Capturing Cross-document Relations for MDS}
\label{sec:cross_document_relation}

Currently, many MDS models still center on simple concatenation of input documents into a flat sequence, ignoring cross-document relations. Unlike SDS, MDS input documents may contain redundant, complementary, or contradictory information \cite{radev2000common}. Discovering cross-document relations, which can assist models to extract salient information, improve the coherence and reduce redundancy of summaries\cite{li2020leveraging}. Research on capturing cross-document relations has begun to gain momentum in the past two years; one of the most widely studied topics is \textsl{graphical models}, which can easily be combined with deep learning based models such as graph neural networks and Transformer models. Several existing works indicate the efficacy of graph-based deep learning models in capturing semantic-rich and syntactic-rich representation and generating high-quality summaries \cite{wang2020heterogeneous, yasunaga2019scisummnet, li2020leveraging, yasunaga2017graph}. To this end, a promising and important direction would be to design a better mechanism to introduce different graph structures \cite{christensen2013towards} or linguistic knowledge \cite{bing2015abstractive, ma2021incorporating}, possibly into the attention mechanism in deep learning based models, to capture cross-document relations and to facilitate summarization.

\subsection{Creating More High-quality Datasets for MDS}

Benchmark datasets allow researchers to  train, evaluate and compare the capabilities of different models on the same stage. High-quality datasets are critical to develop MDS tasks. DUC and TAC, the most common datasets used for MDS tasks, have a relatively small number of samples so are not very suitable for training DNN models. In recent years, some large datasets have been proposed, including  WikiSum \cite{liu2018generating}, Multi-News \cite{fabbri-etal-2019-multi}, and WCEP \cite{demian2020largescale}, but more efforts are still needed.
Datasets with documents of rich diversity, with minimal positional and extractive biases are desperately required to promote and accelerate MDS research, as are datasets for other applications such as summarization of medical records or dialogue \cite{molenaar2020medical}, email \cite{ulrich2008publicly, zajic2008single}, code \cite{rodeghero2014improving, mcburney2014automatic}, software project activities \cite{alghamdi2020human}, legal documents \cite{kanapala2019text}, and multi-modal data \cite{li2020aspect}.
The development of large-scale cross-task datasets will facilitate multi-task learning \cite{xu2020matinf}. However, the datasets of MDS combining with text classification, question answering, or other language tasks have seldom been proposed in the MDS research community, but these datasets are essential and widely employed in industrial applications.

\subsection{Improving Evaluation Metrics for MDS}

To our best knowledge, there are no evaluation metrics specifically designed for MDS models -- SDS and MDS models share the same evaluation metrics. New MDS evaluation metrics should be able to: (1) evaluating the relations between the different input documents in the generated summary; (2) measuring to what extent the redundancy in input documents is reduced; and (3) judging whether the contradictory information across documents is reasonably handled. A good evaluation indicator is able to reflect the true performance of an MDS model and guide design of improved models. However, current evaluation metrics \cite{fabbri2021summeval} still have several obvious defects. For example, despite the effectiveness of commonly used ROUGE metrics, they struggle to accurately measure the semantic similarity between a golden and generated summary because ROUGE-based evaluation metrics only consider vocabulary-level distances; as such, even if a ROUGE score improves, it does not necessarily mean that the summary is of a higher quality and so is not ideal for model training. Recently, some works extend ROUGE along with WordNet \cite{shafieibavani2018graph} or pre-trained LMs \cite{zhang2020bertscore} to alleviate these drawbacks. It is challenging to propose evaluation indicators that can reflect the true quality of generated summaries comprehensively and as semantically as human raters. Another frontline challenge for evaluation metrics research is unsupervised evaluation, being explored by a number of recent studies \cite{sun2019feasibility,gao2020supert}.

\subsection{Reinforcement Learning for MDS}

Reinforcement learning \cite{mnih2016asynchronous} is a cluster of algorithms based on dynamic programming according to the Bellman Equation to deal with sequential decision problems, where state transition dynamics of the environment are provided in advance. Several existing works \cite{paulus2018deep, narayan2018ranking, yao2018deep} model the document summarization task as a sequential decision problem and adopt reinforcement learning to tackle the task. Although deep reinforcement learning for SDS has made great progress, we still face challenges to adapt existing SDS models to MDS, as the latter suffer from a large state, action space, and problems with high redundancy and contradiction \cite{mao2020multi}. Additionally, current summarization methods are based on model-free reinforcement learning algorithms, in which the model is not aware of environment dynamics but continuously explores the environment through simple trial-and-error strategies, so they inevitably suffer from low sampling efficiencies. Nevertheless, the model-based approaches can leverage data more efficiently since they update models upon the prior to the environment. In this case, data-efficient reinforcement learning for MDS could potentially be explored in the future.

\subsection{Pre-trained Language Models for MDS}

In many NLP tasks, the limited labeled corpora are not adequate to train semantic-rich word vectors. Using large-scale, unlabeled, task-agnostic corpora for pre-training can enhance the generalization ability of models and accelerate convergence of networks \cite{peters2018deep, mikolov2013distributed}. At present, pre-trained LMs have led to successes in many deep learning based NLP tasks. Among the reviewed papers \cite{zhong2020extractive, logan2019scoring, li2020leveraging}, multiple works adopt pre-trained LMs for MDS and achieve promising improvements. Applying pre-trained LMs such as BERT \cite{devlin2018bert}, GPT-2 \cite{radford2019language}, GPT-3 \cite{brown2020language}, XLNet \cite{yang2019xlnet}, ALBERT \cite{lan2020albert}, or T5 \cite{colin2020exploring}, and fine-tuning them on a variety of downstream tasks allows the model to achieve faster convergence speed and can improve model performance. MDS requires the model to have a strong ability to process long sequences. It is promising to explore powerful LMs specifically targeting long sequence input characteristics and avoiding quadratic memory growth for self-attention mechanism, such as Longformer \cite{beltagy2020longformer}, REFORMER \cite{kitaev2020reformer}, or Big Bird \cite{zaheer2020big} with pre-trained models. Also, tailor-designed pre-trained LMs for summarization have not been well-explored, e.g., using gap sentences generation is more suitable than using masked language model \cite{Dzhang2019pegasus}. Most MDS methods focus on combining pre-trained LMs in encoder and, as for capturing cross-document relations, applying them in decoder is also a worthwhile direction for research \cite{pasunuru2021efficiently}.

\subsection{Creating Explainable Deep Learning Model for MDS}

Deep learning models can be regarded as black boxes with high non-linearity; it is extremely challenging to understand the detailed transformation inside them. However, an explainable model can reveal how it generates candidate summaries -- to distinguish whether the model has learned the distribution of generating condensed and coherent summaries from multiple documents without bias -- and is thus crucial for model building. Recently, a large amount of researches into explainable models \cite{zhang2018interpretable,rudin2019stop} have proposed easing the non-interpretable concern of deep neural networks, within which model attention plays an especially important role in model interpretation \cite{zhou2016learning,serrano2019attention}. While explainable methods have been intensively researched in NLP \cite{kumar2020nile,jain2020learning}, studies into explainable MDS models are relatively scarce and would benefit from future development.

\subsection{Adversarial Attack and Defense for MDS}

Adversarial examples are strategically modified samples that aim to fool deep neural networks based models. An adversarial example is created via the worst-case perturbation of the input to which a robust DNN model would still assign correct labels, while a vulnerable DNN model would have high confidence in the wrong prediction. The idea of using adversarial examples to examine the robustness of a DNN model originated from research in Computer Vision \cite{corr/SzegedyZSBEGF13} and was introduced in NLP by Jia et al. \cite{emnlp/JiaL17}. An essential purpose for generating adversarial examples for neural networks is to utilize these adversarial examples to enhance the model's robustness \cite{corr/GoodfellowSS14}. Therefore, research on adversarial examples not only helps identify and apply a robust model but also helps to build robust models for different tasks. Following the pioneering work proposed by Jia et al. \cite{emnlp/JiaL17}, many attack methods have been proposed to address this problem in NLP applications \cite{tist/ZhangSAL20} with limited research for MDS \cite{ChengYCZH20}. It is worth filling this gap by exploring existing and developing new, adversarial attacks on the state-of-the-art DNN-based MDS models.

\subsection{Multi-modality for MDS}
Existing multi-modal summarization is based on non-deep learning techniques \cite{ li2017multi, jangra2021multi, jangra2020text, jangra2020multi}, leaving a huge opportunity to exploit deep learning techniques for this task. Multi-modal learning has led to successes in many deep learning tasks, such as Visual Language Navigation \cite{hu2020soft} and Visual Question Answering \cite{antol2015vqa}. Combining MDS with multi-modality has a range of applications: 

\begin{itemize}
\item text + image: generating summaries with pictures and texts for documents with pictures. This kind of multi-modal summary can improve the satisfaction of users \cite{zhu2018msmo}; 

\item text + video: based on the video and its subtitles, generating a concise text summary that describes the main context of video \cite{palaskar2019multimodal}. Movie synopsis is one application; 

\item text + audio: generating short summaries of audio files that people could quickly preview without actually listening to the entire audio recording \cite{erol2003multimodal}. 
\end{itemize}

Deep learning is well-suited for multi-modal tasks \cite{guo2019deep}, as it is able to effectively capture highly nonlinear relationships between images, text or video data. 
Existing MDS models target at dealing with textual data only. Involving richer modalities based on textual data requires models to embrace larger capacity to handle these multi-modal data. The big models such as UNITER \cite{chen2020uniter}, VisualBERT \cite{li2019visualbert} deserve more attention in multi-modality MDS tasks.
However, at present, there is little multi-modal research work based on MDS; this is a promising, but largely under-explored, area where more studies are expected.

\section{Conclusion}
In this article, we have presented the first comprehensive review of the most notable works to date on deep learning based multi-document summarization (MDS). We propose a taxonomy for organizing and clustering existing publications and devise the network design strategies based on the state-of-the-art methods. We also provide an overview of the existing multi-document objective functions, evaluation metrics and datasets, and discuss some of the most pressing open problems and promising future extensions in MDS research. We hope this survey provides readers with a comprehensive understanding of the key aspects of the MDS tasks, clarifies the most notable advances, and sheds light on future studies.
\label{sec: conclusion}

\setlength{\bibsep}{0.5ex}
\bibliographystyle{ACM-Reference-Format}
\begin{spacing}{0.5}

\bibliography{sample-base}
\end{spacing}

\appendix

\end{document}